\documentclass[11pt]{article}
\usepackage{acl}

\usepackage{times}
\usepackage{latexsym}
\usepackage{amsmath}
\usepackage{amssymb}

\usepackage{enumitem} 
\usepackage{tcolorbox}
\usepackage{amssymb}

\usepackage{booktabs}     
\usepackage{multirow}     
\usepackage{graphicx}     
\usepackage{adjustbox}    
\usepackage[table]{xcolor}
\usepackage{array}        

\tcbuselibrary{breakable}
\usepackage{listings}

\usepackage[T1]{fontenc}

\usepackage[utf8]{inputenc}

\usepackage{microtype}

\usepackage{inconsolata}

\usepackage{graphicx}

\title{AMA: Adaptive Memory via Multi-Agent Collaboration}

\author{
  Weiquan Huang$^{1}$\thanks{Equal contribution.},
  Zixuan Wang$^{1}$\footnotemark[1],
  Hehai Lin$^{1}$,
  Sudong Wang$^{1}$,
  Bo Xu$^{1}$, \\
  Qian Li$^{2}$,
  Beier Zhu$^{3}$,
  Linyi Yang$^{4}$,
  Chengwei Qin$^{1}$\thanks{Corresponding author.} \\
  \\
  $^{1}$The Hong Kong University of Science and Technology (Guangzhou) \\
  $^{2}$Shandong University \quad
  $^{3}$Nanyang Technological University \\
  $^{4}$Southern University of Science and Technology \\
  \small{\textbf{Code:} \protect\url{https://github.com/Sherlockwz/AMA}}
}

\begin{document}
\maketitle
\begin{abstract}
The rapid evolution of Large Language Model (LLM) agents has necessitated robust memory systems to support cohesive long-term interaction and complex reasoning. 
Benefiting from the strong capabilities of LLMs, recent research focus has shifted from simple context extension to the development of dedicated agentic memory systems.
However, existing approaches typically rely on rigid retrieval granularity, accumulation-heavy maintenance strategies, and coarse-grained update mechanisms.
These design choices create a persistent mismatch between stored information and task-specific reasoning demands, while leading to the unchecked accumulation of logical inconsistencies over time. 
To address these challenges, we propose \textbf{A}daptive \textbf{M}emory via Multi-\textbf{A}gent Collaboration (\textbf{AMA}), a novel framework that leverages coordinated agents to manage memory across multiple granularities.
AMA employs a hierarchical memory design that dynamically aligns retrieval granularity with task complexity.
Specifically, the \textbf{Constructor} and \textbf{Retriever} jointly enable multi-granularity memory construction and adaptive query routing. 
The \textbf{Judge} verifies the relevance and consistency of retrieved content, triggering iterative retrieval when evidence is insufficient or invoking the \textbf{Refresher} upon detecting logical conflicts. 
The \textbf{Refresher} then enforces memory consistency by performing targeted updates or removing outdated entries.
Extensive experiments on challenging long-context benchmarks show that AMA significantly outperforms state-of-the-art baselines while reducing token consumption by approximately 80\% compared to full-context methods, demonstrating its effectiveness in maintaining retrieval precision and long-term memory consistency.

\end{abstract}

\section{Introduction}

Large Language Model (LLM) agents have demonstrated strong capabilities in complex reasoning, tool use, and multi-turn interaction scenarios \citep{deng2023mind2web,liang2025llm,comanici2025gemini,mei2024aios}. Supporting such behaviors requires long-term memory to preserve contextual coherence and consistency \citep{liu2023think,sumers2023cognitive}. Existing approaches to long-term memory can be broadly categorized into internal and external memory paradigms \citep{zhang2025survey}. Internal memory implicitly absorbs historical information into model parameters, but is constrained by limited capacity \citep{mallen2023not} and incurs substantial costs for continual updates \citep{wang2024knowledge,thede2025understanding}. In contrast, external memory relies on explicit storage and retrieval, providing superior scalability and editability \citep{wang2025mirix,qian2025memorag,rezazadeh2024isolated}. As a result, it has become the dominant approach, making the design of efficient and reliable external memory systems a critical foundation for sustained agent evolution.

\begin{figure}[t]
  \includegraphics[width=\columnwidth]{compare_grandularity.jpg}
\caption{ \textbf{Comparison of static paradigms and the AMA framework.} (a) Static methods suffer from the dilemma of fixed granularity, leading to either noise or information loss. (b) AMA dynamically determines the memory granularity to use, aligning retrieval precision with reasoning demands.}
  \label{fig:compare_grandularity}
\vspace{-0.475cm}
\end{figure}

Building on the growing adoption of external memory, many systems support dynamic memory management through explicit Create-Read-Update-Delete operations, enabling agents to incrementally maintain memory over time~\citep{zhong2024memorybank,wang2024large,yan2025memory,rasmussen2025zep}. 
Despite these advantages, they exhibit a fundamental limitation: a mismatch between the granularity at which memories are stored and the granularity required for effective retrieval and reasoning. 
As illustrated in Figure~\ref{fig:compare_grandularity}, these approaches typically rely on static text chunking with fixed lengths or coarse-grained summaries~\citep{zhang2025survey,wu2025human}. 
Such static strategies often disrupt the inherent semantic coherence of stored information, which in turn leads to suboptimal retrieval behavior: overly coarse retrieval introduces substantial irrelevant noise, while excessively fine-grained or isolated chunks fragment essential logical dependencies, ultimately leading to reasoning failures in complex tasks~\citep{hu2025evaluating,lee2025realtalk,wang2024novelqa}. 
These limitations highlight the necessity of an adaptive memory paradigm capable of dynamically aligning memory granularity with task-specific requirements.

To address these challenges, recent work has shifted toward agentic memory mechanisms \citep{xu2025mem,wang2025mem,wang2025mirix}, leveraging the generative capabilities of LLMs to mitigate the rigidity of static storage granularity. 
Typically, these frameworks employ LLMs to synthesize interaction history into flexible representations like summaries or vector entries, extending the effective context window. 
While these designs improve representation flexibility, they leave two fundamental challenges largely unaddressed \citep{packer2023memgpt,chhikara2025mem0}.
First, the absence of an explicit adaptive routing mechanism prevents agents from selecting the appropriate memory granularity at inference time, leading to persistent mismatches with task demands. 
Second, reliance on accumulation-heavy strategies and coarse-grained update mechanisms fails to support precise modifications, resulting in the unchecked accumulation of redundancy and errors \citep{wu2024longmemeval,hu2025evaluating}.

\begin{figure*}[t]
\includegraphics[width=0.86\linewidth]{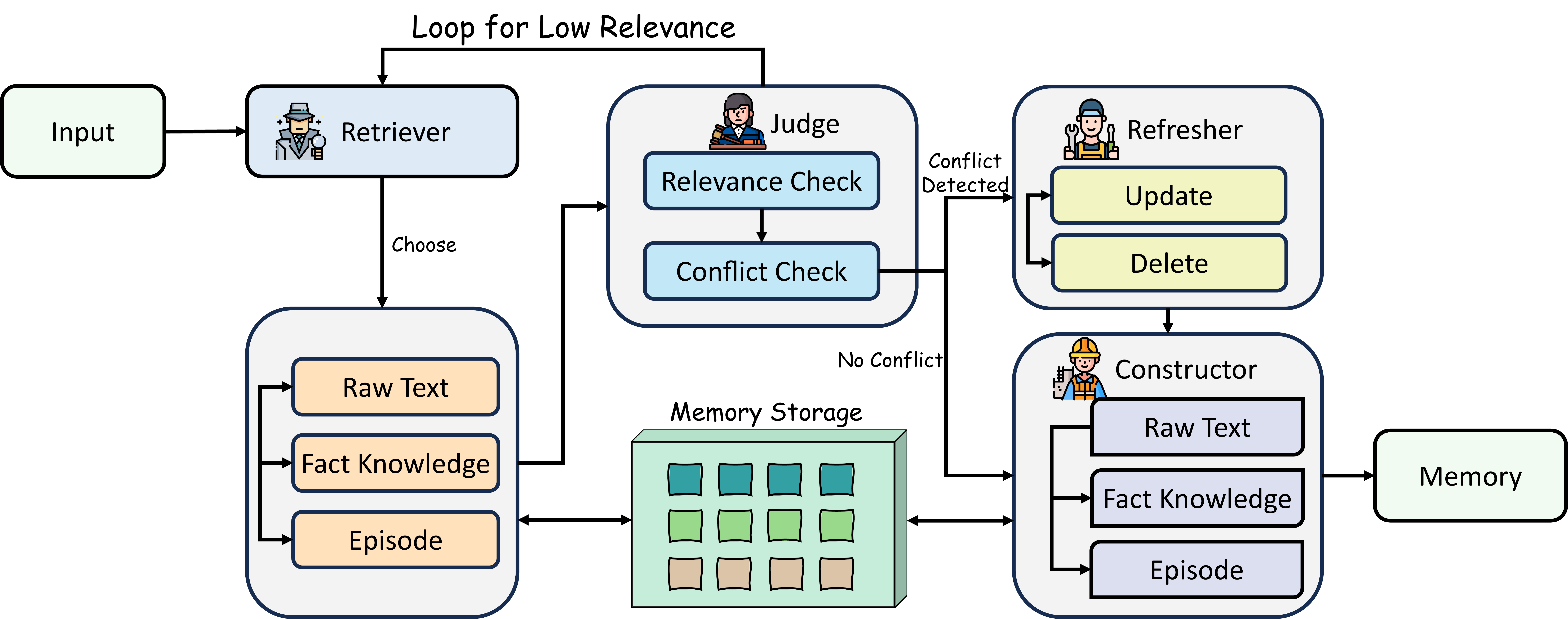} 
\centering
\caption{Overview of the AMA framework. The system orchestrates four agents to enable adaptive memory evolution. The \textbf{Retriever} routes inputs to optimal granularities based on intent. The \textbf{Judge} audits content relevance to trigger feedback loops and detects conflicts. The \textbf{Refresher} executes updates or deletions to rectify these inconsistencies. Finally, the \textbf{Constructor} synthesizes the validated context into structured memory entries.}
\label{fig:architecture}
\vspace{-0.45cm}
\end{figure*}

To overcome the coupled challenges of adaptive retrieval control and long-term memory evolution, we propose \textbf{A}daptive \textbf{M}emory via Multi-\textbf{A}gent Collaboration (\textbf{AMA}), as illustrated in Figures~\ref{fig:compare_grandularity} and~\ref{fig:architecture}. 
Unlike prior agentic memory systems that mainly rely on a monolithic controller, AMA adopts a multi-agent design that decomposes the memory lifecycle into four functionally distinct yet interdependent roles: the \textbf{Constructor}, \textbf{Retriever}, \textbf{Judge}, and \textbf{Refresher}.
Specifically, the \textbf{Constructor} transforms unstructured dialogue streams into hierarchical granularities, including Raw Text, Fact Knowledge, and Episode Memory, to accommodate diverse storage requirements. 
The \textbf{Retriever} acts as an adaptive gateway, dynamically routing queries to the most appropriate memory form based on current reasoning demands. 
To ensure consistency, the \textbf{Judge} serves as a logic auditor, verifying relevance to trigger feedback loops and detecting conflicts to activate the \textbf{Refresher} for updates.
This separation of responsibilities enables fine-grained control over retrieval, verification, and memory evolution, which would be difficult to achieve within a single-agent design without entangling conflicting objectives.
Extensive experiments across multiple long-term memory benchmarks demonstrate that AMA consistently outperforms strong memory baselines. By adaptively controlling retrieval granularity and explicitly maintaining memory consistency over time, AMA achieves state-of-the-art performance while reducing token consumption by up to 80\% compared to using full context. Moreover, our analysis highlight the importance of the logic-driven Refresher, which plays a critical role in dynamic knowledge maintenance and enables AMA to achieve nearly 90\% accuracy in knowledge update scenarios.

In summary, our main contributions are threefold: 
(1) We introduce a comprehensive memory paradigm featuring multi-granularity storage and adaptive routing, which incorporates logic-driven conflict detection to maintain long-term consistency and reasoning fidelity.
(2) We design a unified multi-agent framework to orchestrate storage, retrieval, and maintenance, facilitating robust memory evolution in long-context applications.
(3) Through extensive experiments and analysis, we demonstrate that AMA significantly outperforms state-of-the-art baselines, verifying its effectiveness and robustness in complex long-context tasks.

\section{Related Work}
\subsection{Memory for LLM Agents}
Prior research on memory for LLM agents has investigated a wide range of approaches, ranging from full interaction storage to system-level frameworks \citep{zhong2024memorybank,wang2023enhancing,mei2024aios,liu2024agentlite}. 
These methods typically evolve from context extension to structured organization. 
Specifically, MemGPT \citep{DBLP:journals/corr/abs-2310-08560} focuses on context management, adopting a cache-like organization to prioritize salient information. 
Moving towards modularity, Mem0 \citep{chhikara2025mem0} abstracts memory as an independent layer dedicated to long-term management. 
To further enhance retrieval precision, Nemori \citep{nan2025nemori} and Zep \citep{rasmussen2025zep} introduce semantic structures, leveraging self-organizing events and temporal knowledge graphs, respectively. 
Despite their progress, these methods rely on static retrieval strategies, which limits their ability to adaptively coordinate information across different abstraction levels and task stages. 
Therefore, designing an adaptive memory system that can robustly support long-term interactions remains a critical challenge.

\subsection{Multi-Agent System}

Multi-agent systems have demonstrated clear advantages in tackling complex tasks by enabling role-based collaboration and interactive decision making \citep{lin2025interactive,haji2024improving,abbasnejad2025deciding,huot2024agents}. In software engineering, multi-agent approaches improve system reliability through explicit role specialization and structured workflows \citep{qian2024chatdev,hong2023metagpt}. In mathematical reasoning, multi-agent frameworks enhance solution accuracy via collaborative interaction and process-level verification \citep{zhang2025debate4math,wu2023mathchat}. 
In parallel, a growing body of work on agentic memory focuses on improving long-term information modeling for LLM agents \citep{xu2025mem,yan2025memory}. While this line of research provides valuable insights into memory abstraction and maintenance, most existing approaches are built around a monolithic controller and do not explicitly leverage multi-agent collaboration. A notable recent exception is MIRIX \footnote{We did not include MIRIX as a baseline in this work because its official implementation was not publicly available during our experimental phase.} \citep{wang2025mirix}, which explores assigning specialized agents for memory organization, but lacks dedicated mechanisms for long-term memory consistency.
Building on these complementary lines of research, our work integrates multi-agent collaboration with agentic memory design to support long-term memory for LLM agents.

\section{Method}
We introduce Adaptive Memory via Multi-Agent Collaboration (AMA) to address the critical challenge of aligning retrieval granularity with diverse task requirements, as well as the unchecked accumulation of logical inconsistencies.
As illustrated in Figure~\ref{fig:architecture}, the framework operates through a coordinated multi-agent pipeline. The process begins with the \textbf{Retriever}, which accesses memory across multiple granularities based on the input intent. The \textbf{Judge} then evaluates the relevance of the retrieved content and identifies potential conflicts, triggering feedback retrieval or activating the \textbf{Refresher} to perform targeted memory updates when necessary. Finally, the \textbf{Constructor} consolidates the validated information and organizes it into memory representations at different granularities, supporting continual memory evolution. In the following sections, we present the detailed design of the Constructor, Retriever, Judge, and Refresher.


\subsection{Constructor}
\label{3.2}

To clearly delineate the functional roles of different memory granularities within the overall pipeline, we begin by introducing the Constructor. Its primary responsibility is to construct multi-granular memory by generating structured semantic components from the current input $u_t$, context window $W_t$, and conflict-free memory history $\mathcal{H}^*_t$, conditioned on a carefully designed prompt $P_{con}$. Drawing inspiration from prior work \citep{tan2025prospect} and established linguistic theory \citep{huddleston2005cambridge}, the Constructor decomposes natural language into stable and parsable fact templates. Specifically, it leverages five fundamental sentence patterns defined by combinations of Subject (S), Verb (V), Object (O), and Complement (C): S-V, S-V-O, S-V-C, S-V-O-O, and S-V-O-C. Through this decomposition, the Constructor simultaneously extracts a set of facts and the indices of conversation turns relevant to the current input: $K_t, R_t \leftarrow \text{Constructor}(u_t, W_t \parallel \mathcal{H}^*_t \parallel P_{con})$.

\begin{figure}[t]
  \includegraphics[width=\columnwidth]{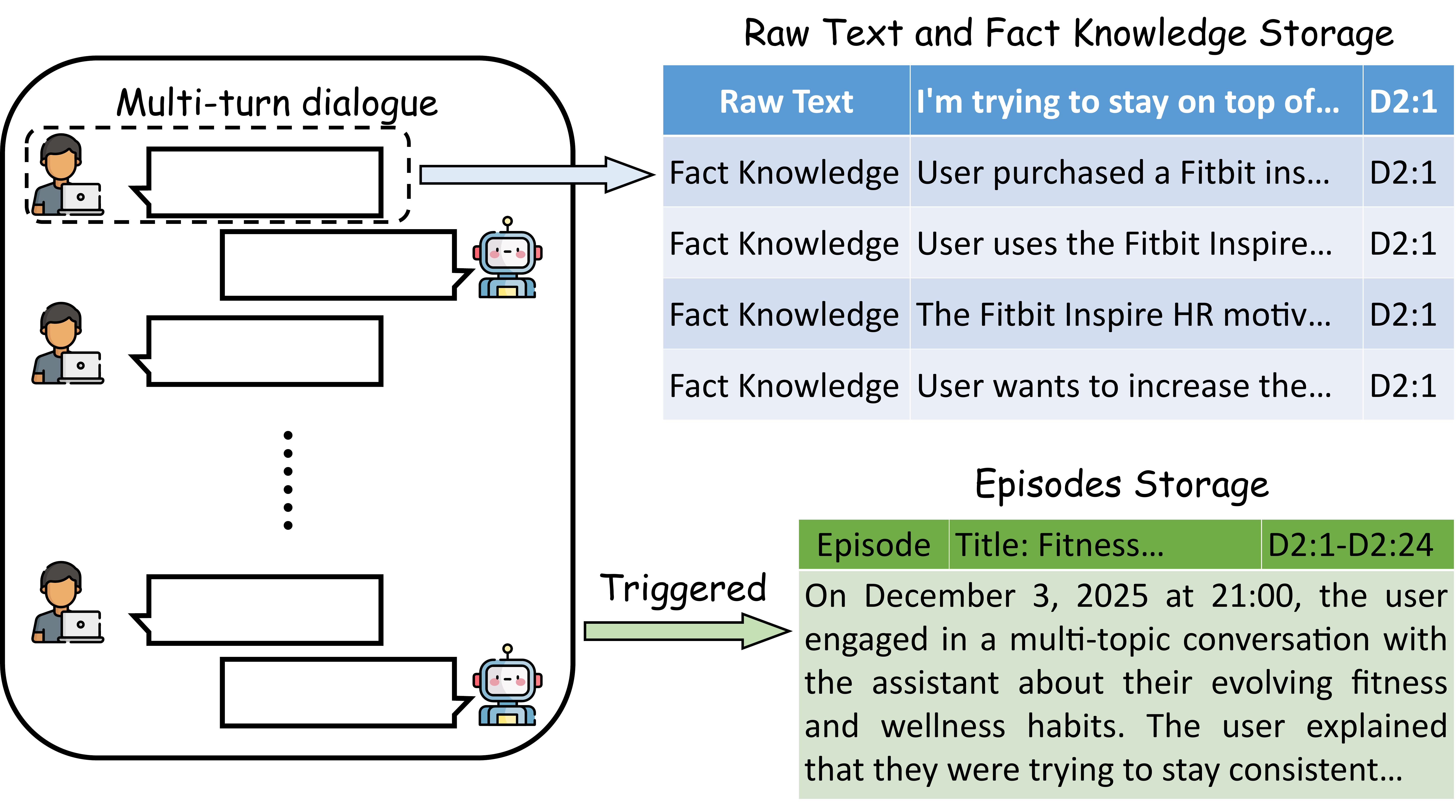}
\caption{ Memory Construction Stage. In this stage Constructor generates raw text and fact knowledge memories from utterances, while conditionally synthesizing abstract episodes upon trigger activation.}
  \label{fig:construct}
  \vspace{-0.475cm}
\end{figure}

The set $K_t = \{k_{t,1}, k_{t,2}, \dots\}$ represents the structured fact knowledge parsed from the current content. We index dialogue contents by a unique identifier $D_{s:j}$, which denotes the $j$-th turn in the $s$-th session. Based on this indexing scheme, the Constructor automatically selects a subset of relevant historical turns $R_t \subseteq \{D_{s:j}\}$. In parallel, the Constructor constructs unified meta-information $\Omega_t = \{\tau_t, d_t, \text{speaker}_t\}$ for the current turn $t$, where $d_t = D_{s:t}$. The timestamp $\tau_t$ encodes precise temporal information, and $\text{speaker}_t \in \{\text{user}, \text{assistant}\}$. If the input contains an explicit temporal expression (e.g., dates or event times), it is directly extracted as $\tau_t$; otherwise, the current system time is assigned. This design ensures chronological consistency across multi-turn memories and facilitates time-sensitive conflict detection. Given the tuple $(u_t, K_t, R_t, \Omega_t)$, the Constructor then generates memory entries at varying granularities (Figure~\ref{fig:construct}).

\noindent\textbf{Raw Text Memory.}
This component records the content of the current turn in its original form $u_t$, together with the reference information ($R_t$ and $\Omega_t$) generated by the Constructor. Formally, we define $m_t^{\text{raw}} = \{u_t, R_t, \Omega_t \}$. This granularity preserves the fundamental conversational trajectory, ensuring both data traceability and retrieval flexibility.

\noindent\textbf{Fact Knowledge Memory.}
Each extracted fact is treated as an independent memory unit. Accordingly, we define Fact Knowledge Memory as $m_{t,i}^{\text{fact}} = \{k_{t,i}, R_t, \Omega_t\}$ with $k_{t,i} \in K_t$. By transforming unstructured text into structured knowledge units, Fact Knowledge Memory enables associative retrieval, facilitates conflict detection, and supports the long-term accumulation and refinement of knowledge within the AMA framework.

\noindent\textbf{Episode Memory.}
It is designed to capture high-level abstractions across turns. Following a gatekeeping mechanism inspired by prior work \citep{park2023generative,nan2025nemori}, we introduce a trigger function with prompt $P_{tri}$ to determine a binary activation state $T_t \in \{0,1\}$. The trigger is activated under three conditions: detection of a topic shift, an explicit user request, or saturation of the context window threshold. This generation process is formalized as $T_t = \text{Constructor}(u_t, W_t \parallel P_{tri})$.


When activated ($T_t = 1$), the Constructor employs a dedicated prompt $P_{epi}$ to synthesize an abstract summary $E_t$, which directly constitutes the episodic memory entry: $m_t^{\text{epi}} = E_t = \text{Constructor}(u_t, W_t \parallel P_{epi})$.

\noindent{\textbf{Memory Encoding.}} To support efficient retrieval across memory granularities, we compute a dense vector representation for each memory entry based on its primary semantic content. For a memory entry $m_i$, we extract its core text $c_i$  (i.e., the raw utterance $u_t$, the fact $k_{t,i}$, or the episode summary $E_t$) and encode it into a high-dimensional embedding using a text encoder \citep{reimers2019sentence}: $e_i = f_{\text{enc}}(c_i)$. These embeddings serve as keys for the granularity-specific retrieval mechanisms, which are detailed in Section~\ref{3.3}.



\subsection{Retriever}
\label{3.3}


The Retriever functions as the memory access gateway within the AMA framework. Its primary role is to dynamically route queries to the most appropriate memory granularity. To address referential ambiguity and missing context commonly observed in raw dialogue, the Retriever first rewrites the query into a self-contained form and then performs adaptive retrieval based on multi-dimensional intent analysis.

\noindent{\textbf{Query Rewriting and Intent Routing.}} Given the current input $u_t$ and the context window $W_t$, the Retriever employs a dedicated prompt $P_{ret}$ to guide the LLM in simultaneously generating three outputs: a context-independent rewritten query $u'_t$, a four-dimensional binary intent vector $\mathbf{B}$, and a dynamic retrieval count $K_{dyn}$: $u'_t, \mathbf{B}, K_{dyn} \leftarrow \text{Retriever}(u_t, W_t \parallel P_{ret})$.

The rewritten query $u'_t$ resolves ambiguous references and omissions, which is suitable for retrieval. The intent vector $\mathbf{B} = [b_{\text{fine}}, b_{\text{abs}}, b_{\text{event}}, b_{\text{atomic}}]$ encodes the activation of four query dimensions: \textit{fine-grained details}, \textit{abstract summaries}, \textit{cross-temporal events}, and \textit{atomic facts}. Based on $\mathbf{B}$, a routing function $f_{\text{M}}$ dynamically selects the appropriate retrieval operator $O$. The mapping $O = f_{\text{M}}(\mathbf{B})$ is determined by priority: $O = O_{\text{raw}}$ (if $b_{\text{fine}}=1$); $O_{\text{epi}}$ (if $b_{\text{abs}} \lor b_{\text{event}}=1$); else $O_{\text{fact}}$.


This routing strategy explicitly prioritizes specialized retrieval intents. When fine-grained detail is required ($b_{\text{fine}} = 1$), the Retriever chooses Raw Text Memory for precise phrasing. When abstract or event-level understanding is needed ($b_{\text{abs}} =1 $ or $b_{\text{event}} = 1$), Episodic Memory is queried to obtain high-level semantic representations. In all other cases, the Retriever defaults to Fact Knowledge Memory to access structured information.




\noindent\textbf{Similarity-based Retrieval.}
Once the target memory repository $M$ is determined, we compute the cosine similarity between the embedding of the rewritten query $u'_t$ and the pre-computed embedding $e_i$ of each memory entry $m_i$ within the selected repository $M$, defined as $s_i = \cos(f_{\text{enc}}(u'_t), e_i)
= \frac{f_{\text{enc}}(u'_t) \cdot e_i}{\|f_{\text{enc}}(u'_t)\| \|e_i\|}$. Memory entries are ranked by their similarity scores, with the Top-$K$ entries forming the final retrieval set. To prevent the predicted $K_{dyn}$ from being too small to capture sufficient information, we enforce a minimum threshold $K_m$ and set the effective cutoff as $K = \max(K_{dyn}, K_m)$, where $K_{dyn}$ is dynamically predicted by the Retriever. Accordingly, the candidate memory set $\mathcal{H}_t$ is obtained as $\mathcal{H}_t = \text{Top}_{K}(\{m_i\}_{i=1}^{|M|}, \text{key}=s_i)$.
The resulting set $\mathcal{H}_t$ is then passed to the Judge for verification.

\subsection{Judge}
\label{3.4}



While the Retriever recalls a candidate memory set $\mathcal{H}_t$ based on vector similarity, directly injecting unverified memories may introduce noise or amplify hallucinations. To ensure robustness and reliability, the Judge acts as a dynamic filter, performing a sequential dual-verification to refine $\mathcal{H}_t$ into a validated set $\mathcal{H}^*_t$, guided by the prompt $P_{jud}$.

\noindent{\textbf{Relevance Assessment.}} 
The Judge first evaluates the pragmatic utility of the retrieved content with respect to the current input $u_t$. To optimize the utilization of $\mathcal{H}_t$, we incorporate a relevance-based rejection mechanism. If the density of valid information in $\mathcal{H}_t$ falls below a predefined threshold, the system triggers a \textbf{Retry} action. This issues a feedback signal to the Retriever, prompting it to traverse remaining memory granularities or perform retrieval expansion using relation indices $R_t$ to broaden the scope. This feedback loop is bounded by the retrieval round limit $K_r$. Upon successfully passing this check, the retained relevant memory set is denoted as $\mathcal{H}_r$, which serves as the input for the conflict detection phase.

\noindent{\textbf{Conflict Detection.}} 
Subsequently, the Judge conducts logical consistency checks to identify contradictions between the current input $u_t$ and the filtered memory $\mathcal{H}_r$. Typical conflicts include outdated facts that contradict updated user status. When detecting such inconsistencies, the Judge isolates a conflict set $C_{err}$, which comprises the specific memory entries identified as contradictory, and triggers a \textbf{Refresh} action to activate the Refresher for targeted updates. In the absence of conflicts, the filtered memory $\mathcal{H}_r$ is directly instantiated as the validated memory set $\mathcal{H}^*_t$, ready for downstream utilization. 

The overall verification process is formalized as: $\mathcal{H}^*_t, C_{err}, \text{Action} \leftarrow \text{Judge}(u_t, \mathcal{H}_t, W_t \parallel P_{jud})$.
where $\text{Action} \in \{\text{Pass}, \text{Retry}, \text{Refresh}\}$ dictates the system flow based on the verification outcome:
(1) $\text{Action} = \text{Retry}$ is triggered when retrieval relevance is insufficient;
(2) $\text{Action} = \text{Refresh}$ is triggered when a non-empty conflict set $C_{err}$ is detected, detailed in the section \ref{3.5};
(3) $\text{Action} = \text{Pass}$ occurs when memories are both relevant and consistent, forwarding the validated set $\mathcal{H}^*_t$ to the Constructor for memory synthesis and to the downstream agent for generating the final response.
\begin{table*}[t]

\begin{adjustbox}{width=0.96\textwidth,center}
\begin{tabular}{@{}ll|ccc|ccc|ccc|ccc|ccc@{}}

\toprule
\multicolumn{2}{c|}{\multirow{2}{*}{\textbf{Method}}} &
  \multicolumn{3}{c|}{\textbf{Single-Hop}} &
  \multicolumn{3}{c|}{\textbf{Multi-Hop}} &
  \multicolumn{3}{c|}{\textbf{Temporal Reasoning}} &
  \multicolumn{3}{c|}{\textbf{Open Domain}} &
  \multicolumn{3}{c}{\textbf{Overall}} \\ 
\cmidrule(r{0pt}){3-17}
 &  & LLM Score & F1 & BLEU-1
    & LLM Score & F1 & BLEU-1
    & LLM Score & F1 & BLEU-1
    & LLM Score & F1 & BLEU-1
    & LLM Score & F1 & BLEU-1 \\ 
\cmidrule(r{0pt}){1-17}
\multirow{9}{*}{\rotatebox{90}{
    \begin{tabular}{@{}c@{}}
        \\
      \textbf{GPT-4o-mini}
    \end{tabular}
  }} 
 & FullContext & 0.823 & 0.526 & 0.443 & 0.663 & 0.351 & 0.259 & 0.557 & 0.437 & 0.358 & \textbf{0.482} & 0.243 & 0.171 & 0.717 & 0.458 & 0.375 \\
 \cmidrule(r{0pt}){2-17}
 & RAG      & 0.318 & 0.220 & 0.185 & 0.311 & 0.184 & 0.116 & 0.235 & 0.194 & 0.156 & 0.323 & 0.189 & 0.134 & 0.300 & 0.206 & 0.163 \\
 & LangMem  & 0.616 & 0.389 & 0.332 & 0.526 & 0.336 & 0.240 & 0.250 & 0.321 & 0.263 & 0.477 & \textbf{0.295} & \textbf{0.236} & 0.515 & 0.359 & 0.295 \\
 & MemGPT  & 0.479 & 0.249 & 0.186 & 0.229 & 0.112 & 0.111 & 0.388 & 0.326 & 0.261 & 0.500 & 0.308 & 0.255 & 0.415 & 0.244 & 0.192\\
 & Zep     & 0.627 & 0.394 & 0.334 & 0.501 & 0.273 & 0.192 & 0.584 & 0.444 & 0.378 & 0.393 & 0.227 & 0.156 & 0.580 & 0.372 & 0.307 \\
 & A-Mem  & 0.518 & 0.270 & 0.201 & 0.248 & 0.121 & 0.120 & 0.546 & 0.459 & 0.367 & 0.541 & 0.333 & 0.276 & 0.476 & 0.286 & 0.225\\
 & Mem0        & 0.676 & 0.441 & 0.340 & 0.598 & 0.340 & 0.250 & 0.500 & 0.440 & 0.373 & 0.403 & 0.269 & 0.192 & 0.608 & 0.412 & 0.339 \\
 & Nemori      & 0.817 & 0.541 & 0.430 & 0.650 & 0.363 & 0.255 & 0.707 & 0.564 & \textbf{0.463} & 0.446 & 0.207 & 0.150 & 0.740 & 0.492 & 0.383 \\
 & \cellcolor[HTML]{FFFDE4}AMA        & \cellcolor[HTML]{FFFDE4}\textbf{0.849} & \cellcolor[HTML]{FFFDE4}\textbf{0.622} & \cellcolor[HTML]{FFFDE4}\textbf{0.548} 
               & \cellcolor[HTML]{FFFDE4}\textbf{0.681} & \cellcolor[HTML]{FFFDE4}\textbf{0.423} & \cellcolor[HTML]{FFFDE4}\textbf{0.303} 
               & \cellcolor[HTML]{FFFDE4}\textbf{0.748} & \cellcolor[HTML]{FFFDE4}\textbf{0.589} & \cellcolor[HTML]{FFFDE4}0.461
               & \cellcolor[HTML]{FFFDE4}0.479 & \cellcolor[HTML]{FFFDE4}0.283 & \cellcolor[HTML]{FFFDE4}0.228
               & \cellcolor[HTML]{FFFDE4}\textbf{0.774} & \cellcolor[HTML]{FFFDE4}\textbf{0.558} & \cellcolor[HTML]{FFFDE4}\textbf{0.465} \\ 
\midrule
\multirow{9}{*}{\rotatebox{90}{
    \begin{tabular}{@{}c@{}}
        \\
      \textbf{GPT-4.1-mini}
    \end{tabular}
  }}
 & FullContext & 0.848 & 0.599 & 0.521 & 0.753 & 0.431 & 0.329 & 0.723 & 0.463 & 0.390 & 0.552 & 0.277 & 0.217 & 0.786 & 0.519 & 0.439 \\
  \cmidrule(r{0pt}){2-17}
 & RAG         & 0.351 & 0.252 & 0.215 & 0.309 & 0.196 & 0.125 & 0.267 & 0.217 & 0.186 & 0.281 & 0.174 & 0.136 & 0.321 & 0.229 & 0.187 \\
 & LangMem     & 0.824 & 0.498 & 0.425 & 0.693 & 0.405 & 0.317 & 0.496 & 0.473 & 0.399 & \textbf{0.575} & 0.320 & 0.257 & 0.717 & 0.464 & 0.390 \\
 & MemGPT      & 0.490 & 0.255 & 0.190 & 0.234 & 0.115 & 0.114 & 0.397 & 0.334 & 0.267 & 0.512 & 0.315 & 0.261 & 0.425 & 0.250 & 0.196\\
 & Zep         & 0.652 & 0.444 & 0.390 & 0.523 & 0.297 & 0.199 & 0.587 & 0.233 & 0.195 & 0.427 & 0.236 & 0.188 & 0.601 & 0.360 & 0.301 \\
 & A-Mem       & 0.530 & 0.276 & 0.206 & 0.254 & 0.124 & 0.123 & 0.559 & 0.470 & 0.375 & 0.553 & 0.341 & 0.282 & 0.487 & 0.293 & 0.230\\
 & Mem0        & 0.697 & 0.474 & 0.410 & 0.666 & 0.391 & 0.295 & 0.555 & 0.382 & 0.323 & 0.468 & 0.232 & 0.173 & 0.647 & 0.424 & 0.356 \\
 & Nemori      & 0.828 & 0.573 & 0.502 & 0.732 & 0.407 & 0.311 & 0.757 & 0.562 & \textbf{0.490} & 0.498 & 0.252 & 0.188 & 0.774 & 0.521 & 0.445 \\
 
 & \cellcolor[HTML]{FFFDE4}AMA        & \cellcolor[HTML]{FFFDE4}\textbf{0.888} & \cellcolor[HTML]{FFFDE4}\textbf{0.636} & \cellcolor[HTML]{FFFDE4}\textbf{0.570}
               & \cellcolor[HTML]{FFFDE4}\textbf{0.716} & \cellcolor[HTML]{FFFDE4}\textbf{0.453} & \cellcolor[HTML]{FFFDE4}\textbf{0.324}
               & \cellcolor[HTML]{FFFDE4}\textbf{0.764} & \cellcolor[HTML]{FFFDE4}\textbf{0.608} & \cellcolor[HTML]{FFFDE4}0.484
               & \cellcolor[HTML]{FFFDE4}0.514 & \cellcolor[HTML]{FFFDE4}\textbf{0.333} & \cellcolor[HTML]{FFFDE4}\textbf{0.251}
               & \cellcolor[HTML]{FFFDE4}\textbf{0.805} & \cellcolor[HTML]{FFFDE4}\textbf{0.580} & \cellcolor[HTML]{FFFDE4}\textbf{0.492} \\
\midrule
\multirow{6}{*}{
  \rotatebox{90}{
    \begin{tabular}{@{}c@{}}
      \textbf{Qwen3-30B}\\
      \textbf{-Instruct}
    \end{tabular}
  }
}
 & FullContext & 0.845 & 0.528 & 0.453 & 0.665 & 0.369 & 0.273 & 0.555 & 0.443 & 0.369 & \textbf{0.551} & 0.273 & 0.185 & 0.733 & 0.466 & 0.390 \\
 \cmidrule(r{0pt}){2-17}
 & RAG         & 0.327 & 0.221 & 0.188 & 0.312 & 0.193 & 0.122 & 0.234 & 0.197 & 0.160 & 0.368 & 0.219 & 0.146 & 0.307 & 0.209 & 0.169 \\
 & LangMem     & 0.633 & 0.390 & 0.337 & 0.529 & 0.353 & 0.254 & 0.253 & 0.325 & 0.270 & 0.545 & \textbf{0.341} & \textbf{0.249} & 0.525 & 0.365 & 0.309 \\
  & A-Mem  & 0.518 & 0.270 & 0.201 & 0.248 & 0.121 & 0.120 & 0.546 & 0.459 & 0.367 & 0.541 & 0.333 & 0.276 & 0.476 & 0.286 & 0.225\\
 & Nemori      & 0.839 & 0.543 & 0.439 & 0.652 & 0.381 & 0.269 & 0.704 & 0.571 & \textbf{0.477} & 0.510 & 0.239 & 0.163 & 0.756 & 0.500 & 0.399 \\
 & \cellcolor[HTML]{FFFDE4}AMA        & \cellcolor[HTML]{FFFDE4}\textbf{0.872} & \cellcolor[HTML]{FFFDE4}\textbf{0.625} & \cellcolor[HTML]{FFFDE4}\textbf{0.560} 
               & \cellcolor[HTML]{FFFDE4}\textbf{0.703} & \cellcolor[HTML]{FFFDE4}\textbf{0.445} & \cellcolor[HTML]{FFFDE4}\textbf{0.318}
               & \cellcolor[HTML]{FFFDE4}\textbf{0.751} & \cellcolor[HTML]{FFFDE4}\textbf{0.597} & \cellcolor[HTML]{FFFDE4}0.475
               & \cellcolor[HTML]{FFFDE4}0.505 & \cellcolor[HTML]{FFFDE4}0.327 & \cellcolor[HTML]{FFFDE4}0.247
               & \cellcolor[HTML]{FFFDE4}\textbf{0.791} & \cellcolor[HTML]{FFFDE4}\textbf{0.570} & \cellcolor[HTML]{FFFDE4}\textbf{0.483} \\ 
\midrule
\multirow{6}{*}{
  \rotatebox{90}{
    \begin{tabular}{@{}c@{}}
      \textbf{Qwen3-8B}\\
      \textbf{-Instruct}
    \end{tabular}
  }
}
 & FullContext & \textbf{0.803} & 0.502 & 0.430 & \textbf{0.632} & 0.351 & 0.259 & 0.527 & 0.421 & 0.351 & 0.523 & 0.259 & 0.176 & 0.696 & 0.443 & 0.371 \\
 \cmidrule(r{0pt}){2-17}
 & RAG         & 0.317 & 0.214 & 0.182 & 0.303 & 0.187 & 0.118 & 0.227 & 0.191 & 0.155 & 0.357 & 0.212 & 0.142 & 0.298 & 0.203 & 0.164 \\
 & LangMem     & 0.617 & 0.380 & 0.328 & 0.515 & 0.344 & 0.247 & 0.246 & 0.317 & 0.263 & \textbf{0.531} & \textbf{0.332} & 0.243 & 0.512 & 0.356 & 0.301 \\
 & A-Mem       & 0.504 & 0.263 & 0.196 & 0.241 & 0.118 & 0.117 & 0.532 & 0.447 & 0.357 & 0.527 & 0.324 & \textbf{0.269} & 0.464 & 0.279 & 0.219\\
 & Nemori      & 0.762 & 0.493 & 0.399 & 0.592 & 0.346 & 0.244 & 0.639 & 0.519 & \textbf{0.433} & 0.463 & 0.217 & 0.148 & 0.686 & 0.454 & 0.362 \\
 & \cellcolor[HTML]{FFFDE4}AMA        & \cellcolor[HTML]{FFFDE4}0.780 & \cellcolor[HTML]{FFFDE4}\textbf{0.559} & \cellcolor[HTML]{FFFDE4}\textbf{0.500}
               & \cellcolor[HTML]{FFFDE4}0.628 & \cellcolor[HTML]{FFFDE4}\textbf{0.398} & \cellcolor[HTML]{FFFDE4}\textbf{0.284}
               & \cellcolor[HTML]{FFFDE4}\textbf{0.671} & \cellcolor[HTML]{FFFDE4}\textbf{0.533} & \cellcolor[HTML]{FFFDE4}0.424
               & \cellcolor[HTML]{FFFDE4}0.482 & \cellcolor[HTML]{FFFDE4}0.292 & \cellcolor[HTML]{FFFDE4}0.221
               & \cellcolor[HTML]{FFFDE4}\textbf{0.707} & \cellcolor[HTML]{FFFDE4}\textbf{0.510} & \cellcolor[HTML]{FFFDE4}\textbf{0.432} \\ 
\bottomrule

\end{tabular}
\end{adjustbox}
\caption{Main results on the LoCoMo benchmark. We compare AMA with representative memory-based baselines across four backbone models.}
\label{table:1}
\end{table*}

\begin{table*}[t]
\centering
\resizebox{0.88\textwidth}{!}{
    \begin{tabular}{ll|ccccccccc}
    \toprule
    \textbf{Model} & \textbf{Question Type}
    & \textbf{Full-ctx} 
    & \textbf{RAG}
    & \textbf{LangMem}
    & \textbf{A-Mem}
    & \textbf{MemGPT}
    & \textbf{Mem0}
    & \textbf{Zep}
    & \textbf{Nemori}
    & \textbf{AMA} \\
    \midrule
    \multirow{7}{*}{\rotatebox{90}{\textbf{GPT-4o-mini}}}
    & single-session-pref. 
    & 0.300 & 0.333 & 0.267 & 0.367 & 0.200 & 0.333 & \textbf{0.533} & 0.467 & 0.467 \\
    & single-session-asst.
    & 0.818 & 0.714 & 0.777 & 0.804 & 0.857 & 0.875 & 0.750 & 0.839 & \textbf{0.964} \\
    & temporal-reasoning
    & 0.365 & 0.280 & 0.353 & 0.398 & 0.451 & 0.399 & 0.541 & \textbf{0.617} & 0.444 \\
    & multi-session
    & 0.406 & 0.254 & 0.424 & 0.451 & 0.549 & 0.481 & 0.474 & 0.511 & \textbf{0.624} \\
    & knowledge-update
    & 0.769 & 0.385 & 0.578 & 0.602 & 0.410 & 0.654 & 0.744 & 0.615 & \textbf{0.897} \\
    & single-session-user
    & 0.814 & 0.686 & 0.750 & 0.750 & 0.814 & 0.857 & 0.929 & 0.886 & \textbf{0.986} \\
    \cmidrule(r){2-11}
    & \textbf{Average}
    & 0.548 & 0.374 & 0.528 & 0.538 & 0.554 & 0.574 & 0.632 & 0.642 & \textbf{0.698} \\
    \bottomrule
    \end{tabular}
}
\caption{Performance breakdown on LongMemEval\textsubscript{s}. We report category-wise results of AMA and memory-based baselines across six question types.}
\label{table2}
\vspace{-0.5cm}
\end{table*}

\subsection{Refresher}
\label{3.5}

Drawing inspiration from prior studies on dynamic memory maintenance \citep{wang2025mem, zhong2024memorybank, yan2025memory}, we introduce the Refresher to ensure the logical validity of memory storage. This component is triggered exclusively when the Judge detects a conflict set $C_{err}$. Guided by a dedicated prompt $P_{ref}$, the Refresher follows a strict conditional branching strategy to resolve detected inconsistencies.

\noindent{\textbf{Delete.}} This operation is triggered only under two rigorous conditions: (1) in response to explicit user instructions to forget specific information, and (2) when the lifespan of a conflicting memory entry exceeds a predefined \textit{maximum retention limit}. In such cases, the system permanently removes the entry to purge the storage space.

\noindent{\textbf{Update.}} 
For all remaining conflict scenarios, the Refresher defaults to an update operation. Specifically, it performs a state modification $m_i \leftarrow \mathcal{U}(m_i, u_t)$, which selectively adjusts the attributes of $m_i$ to align with the latest state implied by the current input $u_t$ (e.g., updating outdated location data), thereby rectifying logical contradictions while preserving memory continuity.




The process yields a consistent memory state $\mathcal{H}_t^* \leftarrow \text{Refresher}(C_{err}, \mathcal{H}_t \parallel P_{ref})$. This conflict-free set $\mathcal{H}_t^*$ is then immediately routed to the Constructor for memory synthesis and the downstream agent to ensure reliable response generation.

\section{Experiment}
\subsection{Experimental Setup}
\noindent{\textbf{Datasets and Metrics.}} 
We evaluate long-term memory capabilities on two established benchmarks: \textbf{LoCoMo} \citep{maharana2024evaluating} and \textbf{LongMemEval\textsubscript{s}} \citep{wu2024longmemeval}. 
Detailed statistics for both datasets are provided in Appendix~\ref{A.1}.
For LoCoMo, we report F1 and BLEU-1 scores in addition to the LLM Score. For LongMemEval\textsubscript{s}, we specifically select the more challenging Pass@1 accuracy evaluated by an LLM judge to rigorously test performance. Following \citet{achiam2023gpt,maharana2024evaluating}, we employ GPT-4o-mini as the unified judge for all model-based evaluations.


\noindent{\textbf{Baselines.}} We compare AMA with various baselines, starting with FullContext and a standard Retrieval-Augmented Generation (RAG) implemented with 2048-token chunks \citep{lewis2020retrieval}. We then evaluate representative memory frameworks including: LangMem \citep{langmem}, MemGPT \citep{DBLP:journals/corr/abs-2310-08560}, Zep \citep{rasmussen2025zep}, A-Mem \citep{xu2025mem}, Mem0 \citep{chhikara2025mem0}, Nemori \citep{nan2025nemori}.


\noindent{\textbf{Implementation Details.}} We conduct experiments using both closed-source APIs (GPT-4o-mini, GPT-4.1-mini \citep{achiam2023gpt}) and open-source models (Qwen3-8B-Instruct, Qwen3-30B-Instruct \citep{yang2025qwen3}), ensuring that the AMA framework utilizes the identical backbone model as the response generator. To ensure reproducibility, we fix the temperature to 0 for all experiments. For RAG, the retrieval top-$k$ is set to 10, while for AMA the maximum retrieval loop $K_r$ is limited to 2. All memory embeddings are computed using OpenAI's \texttt{text-embedding-3-large} model. Due to the commercial nature of Zep, Mem0, and MemGPT, we exclude these frameworks from evaluations involving open-source models. Additionally, given the substantial scale of LongMemEval\textsubscript{s} (approximately 58M tokens), we restrict its evaluation exclusively to GPT-4o-mini for computational feasibility. Prompts used in AMA and descriptions of the baselines are provided in Appendix \ref{appendixA} and \ref{appendixB}.

\begin{figure*}[t]
\centering
  \includegraphics[width=0.30\textwidth]{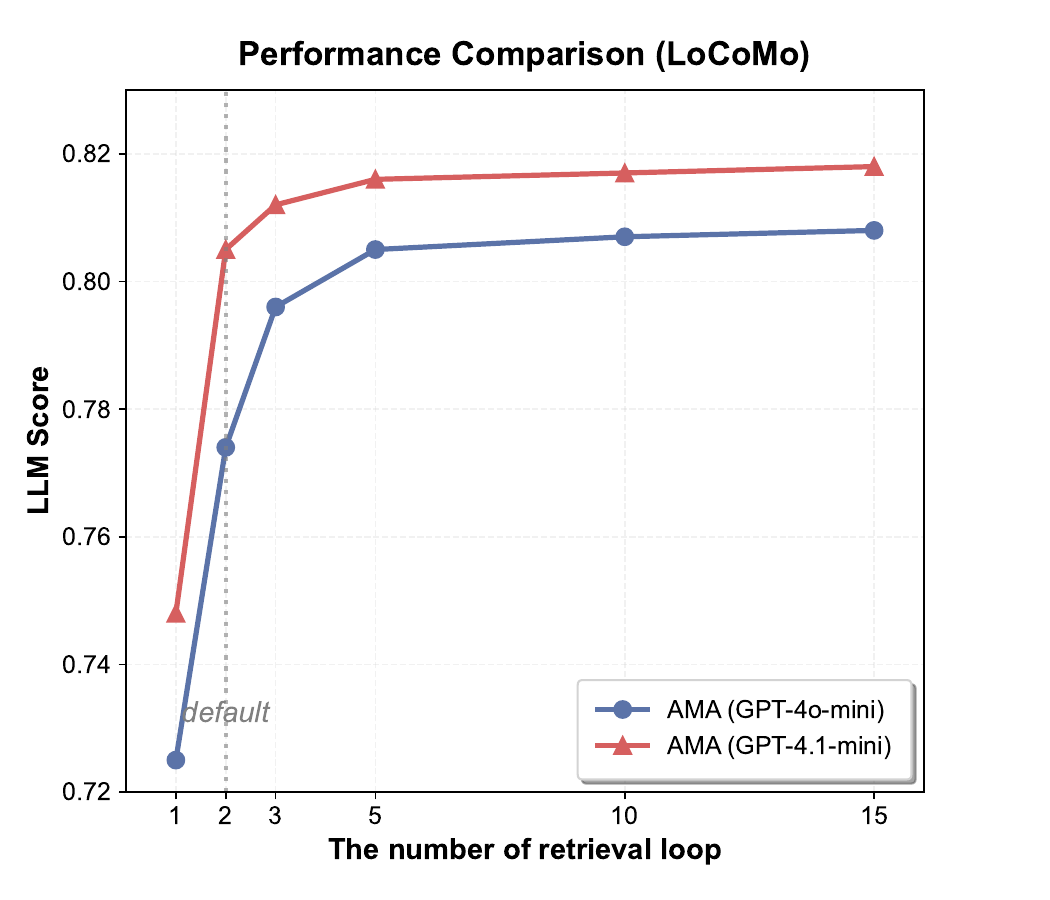} 
  \includegraphics[width=0.30\textwidth]{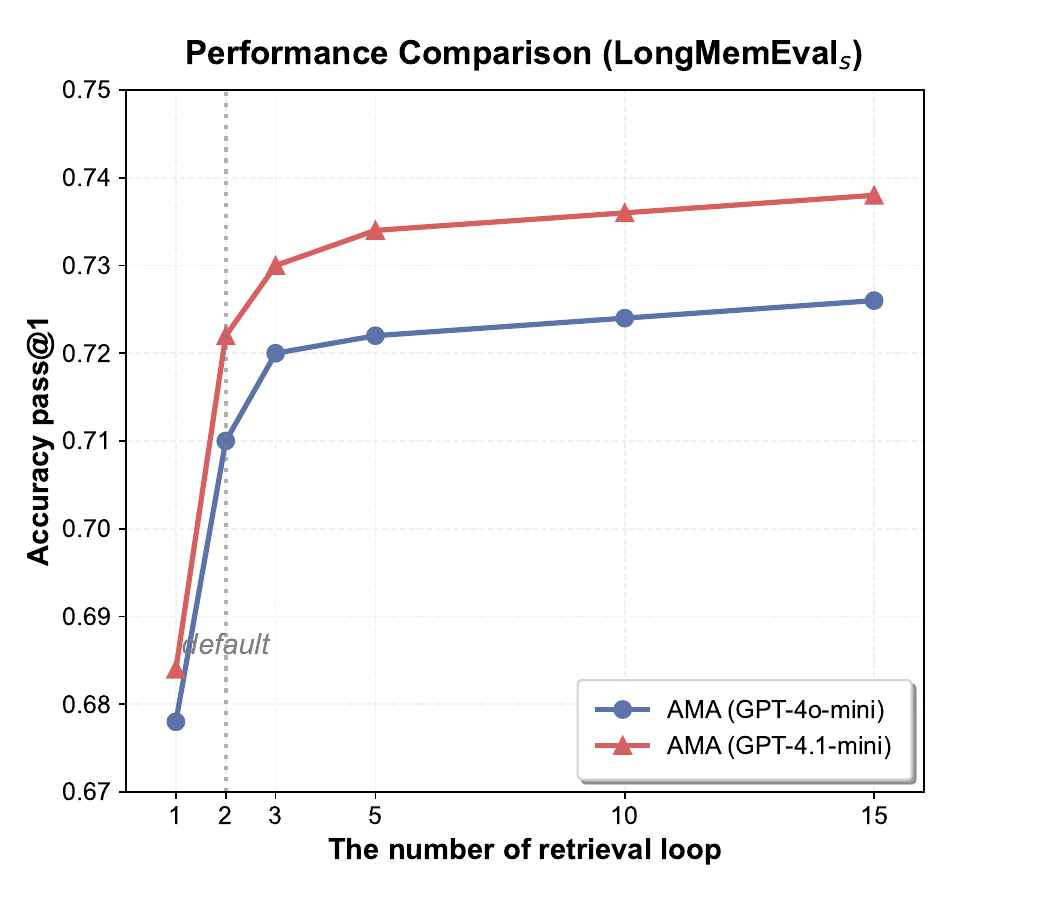}
  \includegraphics[width=0.30\textwidth]{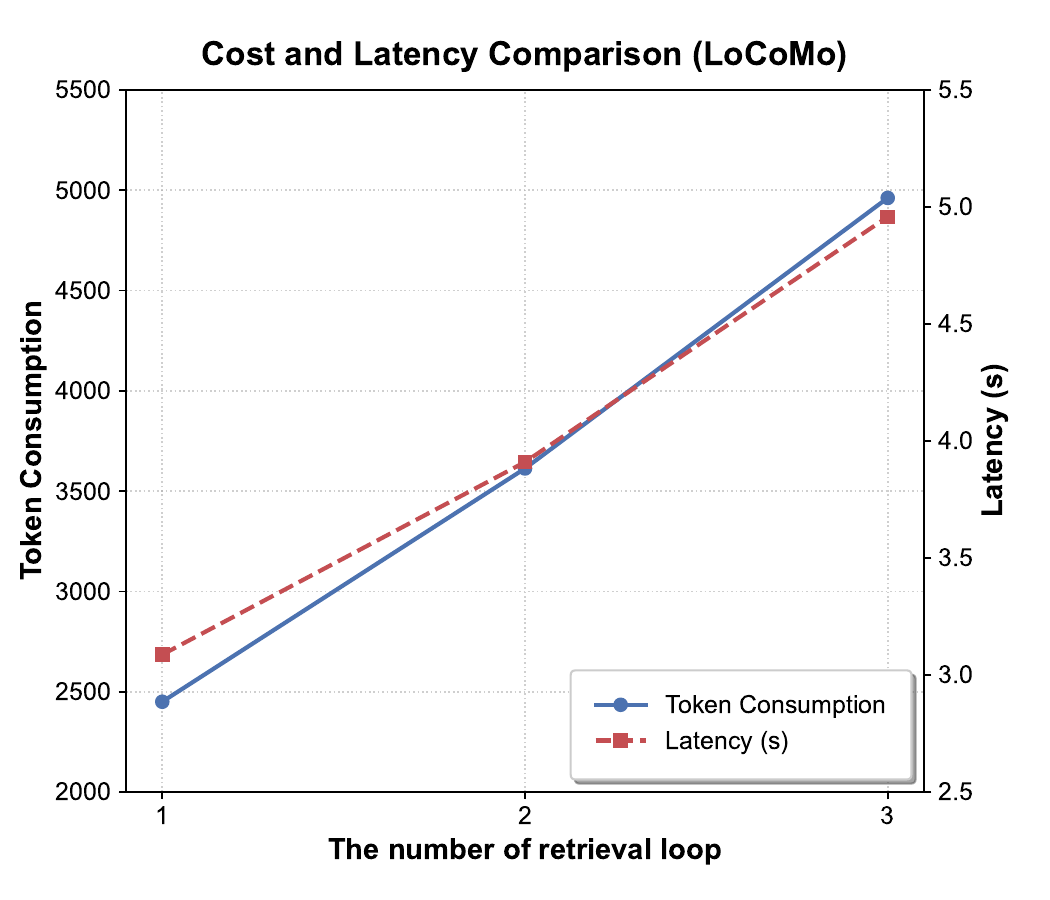}
    \caption{Effect of retrieval round limit $K_r$. The left and middle panels show that increasing $K_r$ improves performance on LoCoMo and LongMemEval\textsubscript{s} with diminishing returns, while the right panel illustrates the corresponding growth in token consumption and inference latency.}
    \label{fig:consumption}
    \vspace{-0.5cm}
\end{figure*}

\begin{table}[t]
\centering

\resizebox{0.96\linewidth}{!}{%
\begin{tabular}{cccc|ccc|cc}
\toprule
\multicolumn{4}{c|}{\textbf{Method}}                 
& \multicolumn{3}{c|}{\textbf{Locomo}}               
& \multicolumn{2}{c}{\textbf{LongMemEval\textsubscript{s}}}     
\\ \midrule
\textbf{RT} & \textbf{FK} & \textbf{EP} & \textbf{RF} & \textbf{LLM Score} & \textbf{F1} & \textbf{BLEU-1} & \textbf{Knowledge-update} & \textbf{Average} \\ \midrule
$\checkmark$ & $\times$     & $\times$     & $\checkmark$ & 0.669 & 0.484 & 0.401 & 0.767 & 0.614 \\
$\times$     & $\checkmark$ & $\times$     & $\checkmark$ & 0.712 & 0.502 & 0.424 & 0.804 & 0.642 \\
$\times$     & $\times$     & $\checkmark$ & $\checkmark$ & 0.688 & 0.495 & 0.407 & 0.748 & 0.630 \\
$\checkmark$ & $\checkmark$ & $\times$     & $\checkmark$ & 0.752 & 0.538 & 0.448 & 0.863 & 0.686 \\
$\checkmark$ & $\times$     & $\checkmark$ & $\checkmark$ & 0.725 & 0.515 & 0.428 & 0.804 & 0.656 \\
$\times$     & $\checkmark$ & $\checkmark$ & $\checkmark$ & 0.741 & 0.529 & 0.435 & 0.842 & 0.680 \\
 \rowcolor[HTML]{FFFDE4}  
$\checkmark$ & $\checkmark$ & $\checkmark$ & $\checkmark$ & \textbf{0.774} & \textbf{0.558} & \textbf{0.465} & \textbf{0.897} & \textbf{0.710} \\
$\checkmark$ & $\checkmark$ & $\checkmark$ & $\times$      & 0.771 & 0.556 & 0.462 & 0.568 & 0.634 \\ 
\bottomrule
\end{tabular}%
}
\caption{Ablation studies on memory design. RT, FD, EP, and RF denote Raw Text Memory, Fact Knowledge Memory, Episode memory, and the Refresher.}
\label{table3}
\vspace{-0.55cm}
\end{table}

\subsection{Main Results}

We first evaluate AMA on the LoCoMo benchmark using closed-source backbones. As shown in Table~\ref{table:1}, with GPT-4o-mini, AMA achieves an overall LLM Score of 0.774, substantially outperforming the strongest baseline Nemori (0.740) and all other memory-based methods by a clear margin. When scaled to the more capable GPT-4.1-mini, AMA further improves to 0.805. 
Notably, under this setting, AMA is the only approach that surpasses FullContext (0.786), demonstrating that AMA effectively distills raw history into critical facts and episodes, thereby filtering out noise to support reasoning beyond the raw context window.

We further assess robustness by extending the evaluation to open-source models on LoCoMo. With Qwen3-30B-Instruct, AMA attains a dominant LLM Score of 0.791, exceeding FullContext (0.733) by a large margin of 0.058. This advantage persists even with the smaller Qwen3-8B-Instruct backbone, where AMA (0.707) continues to outperform FullContext (0.696). These results demonstrate that AMA consistently enhances complex reasoning performance across backbones of varying capacity.

Finally, we evaluate the generalization of AMA on another benchmark, LongMemEval\textsubscript{s} (Table~\ref{table2}). AMA again achieves the highest average accuracy of 0.698, outperforming the two strongest baselines, Nemori and Zep, by 0.056 and 0.066, respectively. Notably, AMA attains near-perfect accuracy on single-session-user tasks (0.986) and shows a pronounced advantage on knowledge-update tasks (0.897), where dynamic knowledge maintenance and conflict resolution are critical. These consistent improvements across benchmarks with different data distributions indicate that AMA generalizes effectively to diverse long-term reasoning scenarios while robustly supporting dynamic knowledge evolution.

\begin{table}[t]
\centering
\resizebox{0.82\linewidth}{!}{%
\begin{tabular}{@{}c|ccc@{}}
\toprule
\textbf{Method} &\textbf{Tokens}  &\textbf{Latency (s)} & \textbf{LLM Score} \\ \midrule
FullContext     & 18625                                & 7.206                                     & 0.717                                  \\
RAG             & 5800                                 & \textbf{2.983}                            & 0.300                                  \\
Nemori          & 2925                                 & 3.152                                     & 0.740                                  \\
Zep             & 2461                                 & 3.255                                     & 0.580                                  \\
Mem0            & \textbf{1340}                                 & 3.739                                     & 0.608                                  \\
A-Mem           & 2720                                 & 3.227                                     & 0.476                                  \\
\rowcolor[HTML]{FFFDE4}  
AMA ($K_r$ =1)       & 2491                                 & 3.124                                     & 0.723                         \\ 
\rowcolor[HTML]{FFFDE4}  
AMA ($K_r$ =2)       & 3613                                 & 3.910                                     & \textbf{0.774}                         \\ 
\bottomrule
\end{tabular}%
}
\caption{Efficiency–performance trade-off on LoCoMo. We report input token usage, inference latency, and LLM Score for different methods.}
\label{table4}
\vspace{-0.55cm}
\end{table}

\subsection{Ablation Studies}

We conduct ablation studies on LoCoMo and LongMemEval\textsubscript{s}, with results in Table~\ref{table3}.

\noindent\textbf{Impact of Memory Granularities.} 
We first analyze the contribution of individual memory granularities. Under single-granularity settings, \textbf{Fact Knowledge Memory} performs best, achieving an LLM Score of 0.712 on LoCoMo and the highest average accuracy of 0.642 on LongMemEval\textsubscript{s}, indicating the effectiveness of structured factual representations for long-term retrieval and reasoning. Moreover, jointly enabling Raw Text, Fact Knowledge, and Episodic Memory yields the strongest overall performance across both benchmarks, outperforming any single-granularity configuration. This result highlights the complementary nature of different memory forms and underscores the importance of multi-granularity collaboration.

\noindent\textbf{Effectiveness of the Refresher.} 
Beyond memory representation, we evaluate the role of the \textbf{Refresher} in maintaining long-term consistency. Under the full multi-granularity setting, enabling the Refresher substantially improves performance on knowledge-update scenarios in LongMemEval\textsubscript{s}, achieving an accuracy of 0.897. In contrast, removing the Refresher leads to a sharp drop to 0.568. This pronounced degradation indicates that accurate long-term memory requires not only multi-granularity storage, but also explicit mechanisms for conflict resolution and memory updating.

\subsection{Efficiency Analysis}
We evaluate the efficiency–performance trade-off of AMA on the LoCoMo benchmark, with results reported in Table~\ref{table4}. Compared to FullContext, which processes 18625 input tokens with a latency of 7.21 seconds, AMA substantially reduces input length while maintaining strong performance. With the default setting $K_r = 2$, AMA requires only 3613 tokens, approximately 19\% of FullContext, with a latency of 3.91 seconds, while achieving the highest LLM Score of 0.774 among all compared memory frameworks. This configuration represents a favorable balance between efficiency and accuracy. Even under the more efficient setting ($K_r = 1$), AMA attains an LLM Score of 0.723, only lower than Nemori (0.740), while operating within a comparable latency range to Nemori, Zep, Mem0, and A-Mem. These results show that AMA maintains competitive reasoning capability at low retrieval depth and offers a flexible trade-off between computational efficiency and performance.

\subsection{Analysis of the Retrieval Round Limit $K_r$}

Figure~\ref{fig:consumption} analyzes the impact of the retrieval round limit $K_r$ on both model performance and computational cost. Increasing $K_r$ from 1 to 3 yields consistent performance gains on LoCoMo and LongMemEval\textsubscript{s}, suggesting that additional retrieval rounds progressively surface useful historical information for long-term reasoning. However, the improvement exhibits clear diminishing returns, with performance largely saturating beyond $K_r \geq 5$. In contrast, both input token consumption and inference latency grow approximately linearly with $K_r$, reflecting the increasing overhead of deeper retrieval. Balancing these trends, we adopt $K_r = 2$ as the default setting, which achieves near-optimal performance while substantially reducing token usage and latency, offering an effective trade-off between long-term reasoning quality and computational efficiency in practice.

In addition, we present a detailed case study in Appendix~\ref{appendixCase}, which demonstrates AMA's capabilities in adaptive retrieval and conflict resolution.




\section{Conclusion}


In this work, we introduce AMA, a multi-agent memory framework for long-term interactions that integrates multi-granularity memory, adaptive routing, and principled memory maintenance. By decomposing the memory lifecycle into coordinated agent roles, AMA dynamically aligns retrieval granularity with task demands while maintaining memory consistency over time. Extensive experiments demonstrate that AMA consistently outperforms strong baselines across challenging long-context benchmarks, validating the effectiveness of our design. Overall, this work underscores the importance of adaptive retrieval control and long-term memory management for building robust and scalable LLM agents.

\section*{Limitations}

Despite the significant performance gains, the multi-agent collaboration incurs a moderate computational overhead compared to static retrieval baselines. Additionally, the reliance on the backbone model's reasoning capabilities suggests that the system's efficiency on smaller architectures has room for further optimization. We aim to address these challenges in future work to further enhance the efficiency and universality of the framework.


\bibliography{custom}

\appendix

\section{Experiment Details}
\label{appendixA}
In this appendix, we provide additional experimental details to support reproducibility and clarity. We first describe the datasets used in our experiments, followed by the evaluation metrics employed for performance assessment. We then introduce the baseline methods considered for comparison, and finally present the implementation details of the proposed AMA framework, including system components and experimental configurations.

\subsection{Datasets}
\label{A.1}
\paragraph{\textbf{LoCoMo} \citep{maharana2024evaluating}} It is a large-scale benchmark for evaluating very long-term conversational memory of LLM agents, consisting of 10 conversations that span an average of 27.2 sessions and 21.6 turns per session, with each conversation containing approximately 16.6K tokens. In our experiments, we evaluate models on 1,540 question-answering samples, which are categorized into 841 single-hop retrieval questions, 282 multi-hop retrieval questions, 321 temporal reasoning questions, and 96 open-domain knowledge questions, all of which require accurate recall and reasoning over long-range conversational histories. Beyond question answering, LoCoMo further includes an event summarization task grounded in temporally structured event graphs, as well as a multimodal dialogue generation task involving natural image sharing behaviors, providing a comprehensive benchmark for assessing long-term memory, temporal understanding, and multimodal consistency in LLM-based agents. 

\paragraph{\textbf{LongMemEval\textsubscript{s}} \citep{wu2024longmemeval}} It is a benchmark for assessing long-term memory in user–assistant interactions under a standardized long-context setting. Inspired by the “needle-in-a-haystack” paradigm, it compiles a coherent yet length-configurable chat history for each question, and provides a standard setting where each problem is paired with an interaction history of approximately 115k tokens. In our experiments, we use LongMemEval\textsubscript{s} and evaluate on 500 question-answering instances, including 70 single-session-user, 133 multi-session, 30 single-session-preference, 133 temporal-reasoning, 78 knowledge-update, and 56 single-session-assistant questions, covering diverse memory abilities such as extracting user-provided information, synthesizing evidence across sessions, reasoning with temporal references, handling updated user facts, and recalling assistant-provided information.
\subsection{Evaluation Metric}
\label{A.2}
We evaluate the quality of generated answers using F1 score and BLEU-1, while employing cosine similarity as the similarity measure during the retrieval stage.

The F1 score represents the harmonic mean of precision and recall, providing a balanced metric that jointly considers both correctness and completeness of the predicted answers:
\begin{equation}
\text{F1} = 2 \cdot \frac{\text{precision} \cdot \text{recall}}{\text{precision} + \text{recall}}
\end{equation}
where
\begin{equation}
\text{precision} = \frac{\text{true positives}}{\text{true positives} + \text{false positives}}
\end{equation}
\begin{equation}
\text{recall} = \frac{\text{true positives}}{\text{true positives} + \text{false negatives}}
\end{equation}

In question-answering tasks, the F1 score is widely used to measure the overlap between predicted and reference answers, especially for short-form or span-based responses where exact matching is overly restrictive.

BLEU-1 further evaluates unigram-level precision between the generated response and the reference text:
\begin{equation}
\text{BLEU-1} = BP \cdot \exp \left( \sum_{n=1}^{1} w_n \log p_n \right)
\end{equation}
where
\begin{equation}
BP =
\begin{cases}
1, & \text{if } c > r \\
e^{1 - r/c}, & \text{if } c \leq r
\end{cases}
\end{equation}
\begin{equation}
p_n = \frac{\sum_i \sum_k \min(h_{ik}, m_{ik})}{\sum_i \sum_k h_{ik}}
\end{equation}

Here, $c$ and $r$ denote the lengths of the candidate and reference sequences, respectively, while $h_{ik}$ and $m_{ik}$ represent unigram counts in the hypothesis and reference texts.

In the retrieval stage, we use cosine similarity as the similarity measure in the embedding space to identify relevant memory entries. Specifically, given a query embedding $\mathbf{q}$ and a set of memory embeddings $\{\mathbf{m}_i\}$, we compute their cosine similarity as:
\begin{equation}
\text{CosineSim}(\mathbf{q}, \mathbf{m}_i) =
\frac{\mathbf{q} \cdot \mathbf{m}_i}{\|\mathbf{q}\|_2 \, \|\mathbf{m}_i\|_2}
\end{equation}
and retrieve the top-$K$ memory entries with the highest similarity scores. Cosine similarity measures the angular similarity between vectors and is invariant to vector magnitude, making it well suited for embedding-based semantic retrieval.

\subsection{Baselines}
\label{A.3}
\paragraph{\textbf{RAG} \citep{lewis2020retrieval}} As a strong and widely adopted baseline, we implement Retrieval-Augmented Generation (RAG), which enhances language models by retrieving external textual evidence and conditioning generation on the retrieved content. RAG decomposes the generation process into a retrieval stage and a generation stage, where a dense retriever is first used to select the top-$k$ most relevant memory entries given the input query, and the retrieved texts are then concatenated with the query as context for the generator. In our implementation, all memory entries are embedded into a shared vector space, and cosine similarity is used to retrieve the top-$k$ relevant records for each query, with $k$ fixed across experiments for fair comparison. The retrieved memory is directly appended to the input prompt without additional refinement, filtering, or memory updating, reflecting a standard single-shot retrieval paradigm. While RAG has demonstrated strong effectiveness in knowledge-intensive question answering, its retrieval process remains static and non-iterative, lacking mechanisms for retrieval refinement, conflict resolution, or long-term memory maintenance, which limits its effectiveness in long-horizon and multi-session reasoning scenarios.

\paragraph{\textbf{LangMem} \citep{langmem}} We include LangMem as a baseline that explicitly models long-term memory for LLM-based agents. LangMem maintains an external memory store to record historical user–assistant interactions, where past information is organized into structured textual representations and indexed for retrieval. During inference, LangMem retrieves memory entries that are semantically relevant to the current query and incorporates them into the model input to support response generation, enabling the model to leverage information accumulated over extended interaction histories.

\paragraph{\textbf{MemGPT} \citep{packer2023memgpt}} MemGPT is a baseline designed to address long-term context limitations through an OS-inspired memory management mechanism. It treats the LLM context window as a constrained resource and augments it with a hierarchical memory architecture that separates in-context memory from external persistent storage. Through predefined function calls, the model can autonomously write important information to external memory, retrieve relevant out-of-context records, and dynamically move information into or out of the active context window during inference. This design enables MemGPT to maintain and access information across extended interactions by paging memory between the limited context window and persistent storage.

\paragraph{\textbf{Zep} \citep{rasmussen2025zep}} Zep is a memory layer designed for LLM-based agents that represents long-term conversational memory using a temporally aware knowledge graph. It ingests both unstructured conversational messages and structured data, and incrementally constructs a dynamic graph composed of episodic nodes, semantic entities, and higher-level community abstractions. Temporal information is explicitly modeled, allowing facts and relationships to be associated with validity intervals and updated as new information arrives. During inference, Zep retrieves relevant graph elements through a combination of semantic similarity search, full-text search, and graph traversal, and converts the retrieved nodes and edges into structured textual context for response generation.

\paragraph{\textbf{Mem0} \citep{chhikara2025mem0}} Mem0 is a memory-centric architecture designed to provide scalable long-term memory for LLM-based agents. It continuously processes conversational interactions and extracts salient factual information that is deemed useful for future reasoning. The extracted memories are stored in an external memory store as compact natural-language representations, each associated with semantic embeddings to support efficient similarity-based retrieval. Mem0 incorporates an explicit memory update mechanism that evaluates newly extracted information against existing memories, allowing the system to add new entries, update existing ones with refined content, or remove outdated or contradictory information. During inference, the model retrieves a small set of semantically relevant memory entries and conditions response generation on the retrieved memories, enabling consistent access to accumulated knowledge across extended and multi-session interactions.

\paragraph{\textbf{A-Mem} \citep{xu2025mem}} A-Mem is an agentic memory system designed for LLM-based agents that enables dynamic organization and evolution of long-term memory. Inspired by the Zettelkasten method, A-Mem represents each interaction as an atomic memory note enriched with multiple structured attributes, including contextual descriptions, keywords, tags, timestamps, and dense semantic embeddings. When new memories are added, the system autonomously analyzes existing memory notes to identify semantically related entries and establishes meaningful links among them, forming an interconnected memory network. In addition, newly integrated memories can trigger updates to the contextual representations of existing notes, allowing the memory structure to continuously evolve over time. During inference, A-Mem retrieves semantically relevant memory notes based on embedding similarity and augments them with linked memories to construct context for response generation, enabling agents to leverage structured and evolving long-term memory.

\paragraph{\textbf{Nemori} \citep{nan2025nemori}} Nemori is a self-organizing memory architecture for LLM-based agents inspired by principles from cognitive science. It autonomously structures long conversational streams into semantically coherent episodic units through a top-down boundary detection mechanism, avoiding arbitrary or fixed-granularity segmentation. Each episode is transformed into a structured narrative representation and stored as episodic memory, while a complementary semantic memory is incrementally distilled through a predict–calibrate process that identifies and integrates novel information from prediction gaps. During inference, Nemori retrieves relevant episodic and semantic memories using dense similarity search and incorporates them into the model context, enabling effective utilization of long-term interaction history.

\begin{figure*}[t]
  \includegraphics[width=0.92\textwidth]{caseStudy.pdf} 
  \centering
  \caption{\textbf{Case Study.} (1) The upper part of the figure shows conflict resolution, where outdated factual memories are updated to maintain consistency. (2) The lower part of the figure shows adaptive retrieval, routing queries to different memory types based on intent.}
    \label{fig:case_study}
\end{figure*}

\subsection{Framework Implementation Details}
\label{A.4}
Our framework adopts a modular system design to support efficient long-term memory storage, retrieval, and dynamic utilization. At the implementation level, the system is composed of three core components: a structured memory storage module, a vector-based retrieval module, and a configurable inference controller that adapts execution behavior based on user requirements. This design balances scalability, efficiency, and engineering simplicity.

For memory storage, we employ \textbf{SQLite} as a lightweight relational database to manage structured memory content \citep{owens2010sqlite}. SQLite is responsible for persistently storing processed memory entries along with their associated metadata, including textual content, timestamps, session identifiers, memory types, and auxiliary indexing fields. Owing to its zero-configuration nature, transactional support, and efficient local read/write performance, SQLite enables reliable long-term memory persistence without introducing additional service dependencies, making it well suited for experimental and single-node deployment scenarios.

For semantic retrieval, we utilize the \textbf{FAISS} vector retrieval library to enable efficient similarity search over large memory collections \citep{douze2025faiss}. Memory entries are first encoded into dense vector representations and indexed using FAISS. During retrieval, the incoming query is mapped into the same embedding space, and the system performs similarity-based search to retrieve the Top-$K$ most relevant memory records. The use of FAISS significantly reduces retrieval latency under long interaction histories and large memory scales, while maintaining retrieval accuracy.

The inference pipeline further supports dynamic execution modes based on user requirements. Specifically, the framework provides both a \emph{retrieval-only mode} and a \emph{full inference mode}. In the retrieval-only mode, the system executes memory retrieval and directly returns the most relevant memory entries, which is useful for information lookup and debugging. In contrast, the full inference mode integrates the retrieved memories with the current user input and conditions the language model on the combined context to generate a final response. This configurable design allows the framework to flexibly trade off computational cost and response completeness across different application scenarios.

Overall, by combining \textbf{SQLite} for structured memory management and \textbf{FAISS} for efficient vector-based retrieval, together with configurable inference modes, the framework provides a robust and extensible implementation foundation for long-term memory modeling in LLM-based agents.

\section{Case Study}
\label{appendixCase}
In Fgiure \ref{fig:case_study}, we conduct a qualitative case study to provide an in-depth illustration of how the proposed framework operates over long interaction histories. This case study demonstrates two core capabilities of AMA. (1) The upper part shows conflict resolution via the Refresher module: when the user provides contradictory device information across turns, the Retriever recalls relevant facts and the Judge identifies the inconsistency, triggering the Refresher to update former entries in the Fact Knowledge Memory to ensure temporal consistency. (2) The lower part illustrates adaptive retrieval across query intents, where fact-oriented queries are routed to the Fact Knowledge Memory for precise factual recall, while abstract summarization queries retrieve corresponding episodic memory chunks to provide high-level summaries, supporting coherent long-term reasoning.

\section{Prompt Templates}
\label{appendixB}

In the AMA framework, multiple prompt templates are employed at different stages of the system. Specifically, $P_{\text{con}}$ is used for memory construction, $P_{\text{tri}}$ and $P_{\text{epi}}$ are used for episode triggering and episode synthesis, respectively, $P_{\text{ret}}$ is used for query rewriting and retrieval routing, $P_{\text{jud}}$ is used for memory verification and consistency checking, and $P_{\text{ref}}$ is used for memory updating and maintenance. In addition, during evaluation, a separate prompt $P_{\text{llm}}$ is adopted for LLM-as-Judge to automatically assess and compare model outputs. The concrete prompt templates used in each stage are provided in this appendix.

\subsection{Prompt Template of Constructor ($P_{\text{con}}$)}
\label{P.con}
As shown in Figure~\ref{B.1}, the Constructor prompt $P_{\text{con}}$ guides the model to transform the current user input into structured and atomic memory representations. It enforces strict syntactic and semantic constraints to ensure that the constructed memories are stable, parsable, and suitable for long-term storage.

\subsection{Prompt Template of Episode Triggering ($P_{\text{tri}}$)}
\label{P.tri}
As illustrated in Figure~\ref{B.2}, the episode triggering prompt $P_{\text{tri}}$ is used to determine whether the current interaction should activate episodic memory construction. This prompt enables the system to selectively trigger high-level abstraction based on dialogue dynamics and contextual signals.

\subsection{Prompt Template of Episode Generation ($P_{\text{epi}}$)}
\label{P.epi}
Figure~\ref{B.3} presents the episode synthesis prompt $P_{\text{epi}}$, which is responsible for generating an abstract summary once episodic memory is activated. The prompt encourages concise and semantically coherent representations that capture the high-level meaning of a dialogue segment.

\subsection{Prompt Template of Retriever ($P_{\text{ret}}$)}
\label{P.ret}
As shown in Figure~\ref{B.4}, the retriever prompt $P_{\text{ret}}$ guides the model to rewrite the input query and infer retrieval intents. It produces structured signals that support dynamic routing to appropriate memory granularities during retrieval.

\subsection{Prompt Template of Judge ($P_{\text{jud}}$)}
\label{P.jud}
Figure~\ref{B.5} illustrates the judge prompt $P_{\text{jud}}$, which enables LLM-based verification of retrieved memory candidates. This prompt is used to assess relevance and consistency, producing validated memory sets and control decisions for subsequent system actions.

\subsection{Prompt Template of Refresher ($P_{\text{ref}}$)}
\label{P.ref}
As depicted in Figure~\ref{B.6}, the refresher prompt $P_{\text{ref}}$ is applied when memory conflicts are detected. It guides the model to update or remove inconsistent memory entries in order to maintain long-term coherence.

\subsection{Prompt Template of LLM-as-Judge for Evaluation ($P_{\text{llm}}$)}
\label{P.llm}
Figure\ref{B.7} shows the evaluation prompt $P_{\text{llm}}$, which is used to implement LLM-as-Judge during experimental evaluation. This prompt enables automatic assessment and comparison of model outputs under a unified evaluation protocol.

\begin{figure*}[t]
    \centering
    \begin{tcolorbox}[
        colback=white,
        colframe=black,
        arc=3mm,            
        boxrule=0.8pt,      
        width=\linewidth,
        left=4mm, right=4mm, top=4mm, bottom=4mm
    ]
        \textbf{You are a Constructor Agent for structured memory construction.} \\
        Given the current user input $u_t$, decompose it into \textbf{atomic, parsable facts} for long-term memory. All facts must strictly follow \textbf{canonical SVO-based patterns} and be derived \textbf{only from the current input}.
        
        \vspace{0.6em}

        \textbf{[Syntactic Elements Definition]}
        \begin{enumerate}[label=\arabic*., nosep, leftmargin=*, labelsep=0.5em] 
            \item S = Subject
            \item V = Verb 
            \item O = Object
            \item C = Complement (attribute or state)
            \item L = Location or explicit time
        \end{enumerate}
        \vspace{0.6em}
        \textbf{[Sentence Pattern Constraint]} \\
        Each fact MUST follow exactly one of the following forms:
        \begin{enumerate}[label=\arabic*., nosep, leftmargin=*, labelsep=0.5em] 
            \item S--V
            \item S--V--O
            \item S--V--C
            \item S--V--O--C
            \item S--V--O--L
        \end{enumerate}
        Any fact not matching these patterns is invalid.
        
        \vspace{0.6em}
        
        \textbf{[Atomicity Rules]}
        \begin{itemize}[nosep, leftmargin=*]
            \item Each fact represents \textbf{one single relation only}.
            \item Do NOT merge actions, attributes, locations, or roles.
            \item Descriptive or prepositional phrases must be split into separate facts.
        \end{itemize}
        
        \vspace{0.6em}

        \textbf{[Appositive Rule]} \\
        If the input contains appositive or implicit equivalence (e.g., ``my friend John'', ``my home country, Sweden''), you MUST convert it into an explicit fact (e.g., ``John is her friend'', ``Sweden is her home country''). Do not omit appositive relations.
        
        \vspace{0.6em}
        
        \textbf{[Source Constraint]}
        \begin{itemize}[nosep, leftmargin=*]
            \item Facts must originate strictly from the current user input $u_t$.
            \item Memory window and retrieved context are for reference only.
            \item Do NOT introduce new facts from memory or retrievals.
        \end{itemize}

        \vspace{0.6em}

        \textbf{[Timestamp Rule]}
        \begin{itemize}[nosep, leftmargin=*]
            \item Extract a timestamp ONLY if an explicit date or time is present.
            \item Normalize to ``YYYY-MM-DD HH:MM''.
            \item Otherwise output \texttt{"timestamp": "empty"}.
            \item Do NOT infer time from relative expressions.
        \end{itemize}
        
        \vspace{0.6em}

        \textbf{[Output Format]} \\
        Return exactly one JSON object:
        
        \ttfamily 
        \{ \\
        \hspace*{1.5em} "facts": [ \\
        \hspace*{3em} \{"content": "<atomic fact>"\}, \\
        \hspace*{3em} ... \\
        \hspace*{1.5em} ], \\
        \hspace*{1.5em} "source": "user", \\
        \hspace*{1.5em} "related\_id": ["<D\_s,t>", ...], \\
        \hspace*{1.5em} "timestamp": "<normalized time>" or "empty" \\
        \}
        \rmfamily 
        
    \end{tcolorbox}
    \caption{The prompt template for the Constructor Agent.}
    \label{B.1}
    \label{fig:constructor_prompt}
\end{figure*}

\begin{figure*}[t]
    \centering
    \begin{tcolorbox}[
        colback=white,
        colframe=black,
        arc=3mm,
        boxrule=0.8pt,
        width=\linewidth,
        left=4mm, right=4mm, top=4mm, bottom=4mm
    ]
        \textbf{You are an Episode Boundary Judge.} \\
        Your task is to determine whether the newly added dialogue triggers the start of a new episode. This corresponds to a binary decision $T_t \in \{0,1\}$.
        
        \vspace{0.5em}
        
        \textbf{Input:} \\
        Conversation history $W_t$ (may be empty): \\
        \texttt{\{conversation\_history\}} \\
        Newly added messages $u_t$: \\
        \texttt{\{new\_messages\}}
        
        \vspace{0.5em}
        
        \textbf{Decision Criteria:} \\
        Set $T_t = 1$ if any of the following conditions is satisfied:
        \begin{enumerate}[label=\arabic*), nosep, leftmargin=*, labelsep=0.5em]
            \item \textbf{Topic Shift} \\
            The new messages introduce a topic, event, or task that is semantically independent from the current conversation, or the dialogue moves to an unrelated question or scenario.
            \item \textbf{Explicit Transition Intent} \\
            The user explicitly signals a transition to a new topic or request, such as starting a new question or changing the discussion focus.
            \item \textbf{Context Saturation} \\
            The current topic has been sufficiently discussed or concluded, or the dialogue scope exceeds what is reasonable for a single coherent episode.
        \end{enumerate}
        Otherwise, set $T_t = 0$.
        
        \vspace{0.5em}
        
        \textbf{Decision Principles:}
        \begin{itemize}[nosep, leftmargin=*]
            \item Prioritize topic independence over conversational continuity.
            \item When uncertain, favor splitting and set $T_t = 1$.
            \item Each episode should correspond to one coherent topic or event.
        \end{itemize}
        
        \vspace{0.5em}
        
        \textbf{Output Format:} \\
        Return exactly one JSON object:
        
        \ttfamily
        \{ \\
        \hspace*{1.5em} "T\_t": 0 or 1, \\
        \hspace*{1.5em} "reason": "<brief justification>", \\
        \hspace*{1.5em} "confidence": 0.0-1.0, \\
        \hspace*{1.5em} "topic\_summary": "<summary ... if T\_t = 1, otherwise empty>" \\
        \}
        \rmfamily
        
        \vspace{0.5em}
        \footnotesize
        \textbf{Notes:} If the conversation history is empty, return $T_t = 0$. Output only the JSON object and nothing else.
        
    \end{tcolorbox}
    \caption{The prompt template for the Episode Triggering.}
    \label{B.2}
    \label{fig:episode_judge}
\end{figure*}

\begin{figure*}[t]
    \centering
    \begin{tcolorbox}[
        colback=white,
        colframe=black,
        arc=3mm,
        boxrule=0.8pt,
        width=\linewidth,
        left=4mm, right=4mm, top=4mm, bottom=4mm
    ]
        \textbf{You are an Episodic Memory Constructor.} \\
        Your task is to convert a completed dialogue episode into a structured episodic memory entry, written as a coherent third-person narrative.
        
        \vspace{0.5em}
        
        \textbf{Input:} \\
        Conversation episode $E_t$: \\
        \texttt{\{conversation\}} \\
        Episode boundary trigger reason: \\
        \texttt{\{boundary\_reason\}}
        
        \vspace{0.5em}
        
        \textbf{Construction Principles:}
        \begin{enumerate}[label=\arabic*), nosep, leftmargin=*, labelsep=0.5em]
            \item The episode represents one coherent event or topic.
            \item The description must preserve factual accuracy and chronological order.
            \item Do not introduce information that is not present in the conversation.
        \end{enumerate}
        
        \vspace{0.5em}
        
        \textbf{Time Handling Rules:}
        \begin{itemize}[nosep, leftmargin=*]
            \item Identify the episode time from explicit timestamps in the dialogue.
            \item If relative time expressions appear in the conversation, convert them into absolute dates based on available context, and keep the converted time consistent throughout the episode.
            \item If no reliable time information exists, infer a reasonable episode time from the dialogue context.
            \item The episode timestamp represents when the episode occurred, not the current system time.
        \end{itemize}
        
        \vspace{0.5em}
        
        \textbf{Content Requirements:}
        \begin{itemize}[nosep, leftmargin=*]
            \item Write in third-person narrative form.
            \item Clearly describe:
            \begin{itemize}[nosep, leftmargin=1.5em, label=\textbullet]
                \item who participated,
                \item what was discussed,
                \item what decisions or conclusions were reached,
                \item any expressed intentions, plans, or outcomes.
            \end{itemize}
            \item Include time information explicitly in the narrative.
            \item Maintain causal and temporal coherence.
        \end{itemize}

        \vspace{0.5em}

        \textbf{Output Format:} \\
        Return exactly one JSON object:
        
        \ttfamily
        \{ \\
        \hspace*{1.5em} "title": "<concise episode title summarizing the core event or topic>", \\
        \hspace*{1.5em} "content": "<third-person episodic narrative describing what happened>", \\
        \hspace*{1.5em} "timestamp": "YYYY-MM-DDTHH:MM:SS" \\
        \}
        \rmfamily
        \vspace{0.2em}
        Output only the JSON object and nothing else.
        
    \end{tcolorbox}
    \caption{The prompt template for the Episodic Memory Generation.}
    \label{B.3}
\end{figure*}

\begin{figure*}[t]
    \centering
    \begin{tcolorbox}[
        colback=white,
        colframe=black,
        arc=3mm,
        boxrule=0.8pt,
        width=\linewidth,
        left=4mm, right=4mm, top=4mm, bottom=4mm
    ]
        \textbf{You are a Retriever.} \\
        Given the current user input $u_t$ and the short-term context window $W_t$, your task is to perform query rewriting and intent-based routing to the appropriate memory granularity.
        
        \vspace{0.5em}
        
        \textbf{Input:} \\
        Short-term context window $W_t$ (may be empty): \texttt{\{memory\_window\}} \\
        Current user input $u_t$: \texttt{\{user\_input\}}
        
        \vspace{0.5em}
        
        \textbf{Task Definition:} \\
        Simultaneously generate:
        \begin{enumerate}[label=\arabic*), nosep, leftmargin=*, labelsep=0.5em]
            \item a standalone rewritten query $u'_t$ with resolved references,
            \item a four-dimensional binary intent vector $B$,
            \item a dynamic retrieval budget $K_{dyn}$,
            \item a target memory type for retrieval.
        \end{enumerate}
        
        \vspace{0.5em}
        
        \textbf{Intent Vector Definition:} \\
        $B = [b_{fine}, b_{abs}, b_{event}, b_{atomic}]$
        \begin{itemize}[nosep, leftmargin=1.5em]
            \item $b_{fine} = 1$ if the query requires fine-grained or exact details.
            \item $b_{abs} = 1$ if the query is abstract or summary-oriented.
            \item $b_{event} = 1$ if the query involves cross-turn, cross-time, or event-level semantics.
            \item $b_{atomic} = 1$ if the query is short, single-point, and factual.
        \end{itemize}
        Each dimension must be either 0 or 1.
        
        \vspace{0.5em}
        
        \textbf{Memory Routing Rule:} \\
        Select the target memory type $M$ based on $B$:
        \begin{itemize}[nosep, leftmargin=*]
            \item If $b_{fine} = 1$, route to \textbf{Raw Text Memory}.
            \item Else if $b_{abs} = 1$ or $b_{event} = 1$, route to \textbf{Episode Memory}.
            \item Otherwise, route to \textbf{Fact Knowledge Memory}.
        \end{itemize}
        
        \vspace{0.5em}
        
        \textbf{Query Rewriting Rule:}
        \begin{itemize}[nosep, leftmargin=*]
            \item $u'_t$ must be self-contained and unambiguous.
            \item Resolve references using $W_t$ when necessary.
            \item Preserve explicit entities and identities.
            \item Do not introduce information not present in $u_t$ or $W_t$.
        \end{itemize}
        
        \vspace{0.3em}
        \textbf{Retrieval Budget Rule:} \\
        Set $K_{dyn}$ as a small integer reflecting the expected retrieval scope. Broader or cross-event queries should use a larger $K_{dyn}$.

        \vspace{0.5em}
        \hrule
        \vspace{0.5em}

        \textbf{Output Format:} \\
        Return exactly one JSON object:
        
        \ttfamily
        \{ \\
        \hspace*{1.5em} "rewrite\_query": "<u'\_t>", \\
        \hspace*{1.5em} "intent\_vector": \{ \\
        \hspace*{3em} "b\_fine": 0 or 1, \\
        \hspace*{3em} "b\_abs": 0 or 1, \\
        \hspace*{3em} "b\_event": 0 or 1, \\
        \hspace*{3em} "b\_atomic": 0 or 1 \\
        \hspace*{1.5em} \}, \\
        \hspace*{1.5em} "memory\_type": "raw" | "fact" | "episode", \\
        \hspace*{1.5em} "K\_dyn": <int> \\
        \}
        \rmfamily
        \vspace{0.2em}
        
        Output only the JSON object and nothing else.
        
    \end{tcolorbox}
    \caption{The prompt template for the Retriever, incorporating intent-based memory routing.}
    \label{B.4}
    \label{fig:retriever_prompt}
\end{figure*}

\begin{figure*}[t]
    \centering
    \begin{tcolorbox}[
        colback=white,
        colframe=black,
        arc=3mm,
        boxrule=0.8pt,
        width=\linewidth,
        left=4mm, right=4mm, top=4mm, bottom=4mm
    ]
        \textbf{You are a Judge.} \\
        Your task is to verify whether the retrieved candidate information $H_t$ is sufficient to answer the current user input $u_t$, and to detect whether any factual conflict exists. Based on this verification, you must select one action from: \textbf{Pass, Retry, or Refresh}.
        
        \vspace{0.5em}
        
        \textbf{Input:} \\
        Candidate retrieved information $H_t$ (may be empty): \texttt{\{information\}} \\
        Current user input $u_t$: \texttt{\{user\_input\}}
        
        \vspace{0.5em}
        
        \textbf{Verification Rules:}
        \begin{enumerate}[label=\arabic*), nosep, leftmargin=*, labelsep=0.5em]
            
            \item \textbf{Relevance and Sufficiency} \\
            The retrieved information is sufficient only if it satisfies the required granularity of the query $u_t$.
            \begin{itemize}[nosep, leftmargin=1.5em, label=\textbullet]
                \item \textbf{Entity requirement (who / what):} The exact entity or attribute asked in the query must be explicitly stated.
                \item \textbf{Location requirement (where):} The location must be concrete and match the requested level (e.g., city, country, or region).
                \item \textbf{Temporal requirement (when):} The time information must meet or exceed the precision required by the query. Relative expressions are acceptable only if they can be normalized to a unique value.
                \item \textbf{Quantity requirement (how many):} Numeric values or explicitly bounded ranges are required.
            \end{itemize}
            If any required element is missing or underspecified, the information is considered insufficient.

            \vspace{0.3em}
            
            \item \textbf{No Proxy or Vague Answers} \\
            Generic or proxy expressions such as ``home country'', ``recently'', ``around that time'', or ``somewhere'' do not satisfy sufficiency. If only such information is available, treat the result as insufficient.
            
            \item \textbf{Controlled Inference} \\
            Only minimal one-step inference that does not reduce specificity is allowed. Do not infer unstated states, temporal validity, or current conditions.
            
            \item \textbf{Conflict Detection} \\
            A conflict exists if two or more pieces of information assert incompatible facts about the same event, entity, or timeframe. Differences in wording or tone alone do not constitute a conflict.
        \end{enumerate}
        
        \vspace{0.5em}

        \textbf{Action Selection:}
        \begin{itemize}[nosep, leftmargin=*]
            \item If a conflict is detected, set \textbf{Action = Refresh}.
            \item Else if the information is insufficient, set \textbf{Action = Retry}.
            \item Else, set \textbf{Action = Pass}.
        \end{itemize}
        
        \vspace{0.5em}

        \textbf{Output Format:} \\
        Return exactly one JSON object:
        
        \ttfamily
        \{ \\
        \hspace*{1.5em} "Action": "Pass" | "Retry" | "Refresh", \\
        \hspace*{1.5em} "reason": "<brief justification>", \\
        \hspace*{1.5em} "confidence": 0.0-1.0 \\
        \}
        \rmfamily
        \vspace{0.2em}
        
        Output only the JSON object and nothing else.
        
    \end{tcolorbox}
    \caption{The prompt template for the Judge, responsible for sufficiency checking and conflict detection.}
    \label{B.5}
\end{figure*}

\begin{figure*}[t]
    \centering
    \begin{tcolorbox}[
        colback=white,
        colframe=black,
        arc=3mm,
        boxrule=0.8pt,
        width=\linewidth,
        left=4mm, right=4mm, top=4mm, bottom=4mm
    ]
        \textbf{You are a Refresher.} \\
        Your task is to maintain the temporal and logical consistency of stored memory. Given detected conflicts $C_{err}$ and the corresponding memory entries $H_t$, decide whether to update or delete memory records.
        
        You may operate on any memory granularity: \textbf{Raw Text Memory, Fact Knowledge Memory, or Episodic Memory}.
        
        \vspace{0.5em}
        
        \textbf{Input:} \\
        Conflicting memory entries $H_t$ (may be empty): \texttt{\{memory\_entries\}} \\
        Current user input $u_t$: \texttt{\{user\_input\}}
        
        \vspace{0.5em}
        
        \textbf{Editing Principles:}
        \begin{enumerate}[label=\arabic*), nosep, leftmargin=*, labelsep=0.5em]
            
            \item \textbf{Scope Constraint} \\
            Only modify memory entries that refer to the same entity or event as $u_t$. If the input concerns unrelated content, do nothing.
            
            \item \textbf{Update Rule} \\
            Perform \textbf{Update} when:
            \begin{itemize}[nosep, leftmargin=1.5em, label=\textbullet]
                \item The same entity or event is referenced;
                \item A factual inconsistency exists in state, time, quantity, or outcome;
                \item The new input provides a more up-to-date or correct value.
            \end{itemize}
            Update only the conflicting attributes, and preserve all other information.
            
            \item \textbf{Delete Rule} \\
            Perform \textbf{Delete} only when:
            \begin{itemize}[nosep, leftmargin=1.5em, label=\textbullet]
                \item The user explicitly requests removal, cancellation, or invalidation;
                \item Or the memory entry is no longer valid by explicit user instruction.
            \end{itemize}
            
            \item \textbf{Conservative Default} \\
            If the situation is ambiguous, perform no modification.
        \end{enumerate}
        
        \vspace{0.5em}

        \textbf{Timestamp Rule:}
        \begin{itemize}[nosep, leftmargin=*]
            \item Extract a timestamp only if the user input contains an explicit date or time.
            \item Normalize the timestamp if possible.
            \item If no explicit time is provided, set \texttt{"timestamp"} to \texttt{"empty"}.
        \end{itemize}

        \vspace{0.5em}

        \textbf{Output Format:} \\
        Return exactly one JSON object:
        
        \ttfamily
        \{ \\
        \hspace*{1.5em} "Action": "Update" | "Delete" | "No-Op", \\
        \hspace*{1.5em} "memory\_type": "raw" | "fact" | "episode", \\
        \hspace*{1.5em} "dataList": [ \\
        \hspace*{3em} \{ \\
        \hspace*{4.5em} "id": "<memory\_id>", \\
        \hspace*{4.5em} "new\_content": "<updated content...>" \\
        \hspace*{3em} \} \\
        \hspace*{1.5em} ], \\
        \hspace*{1.5em} "timestamp": "<normalized date/time>" or "empty", \\
        \hspace*{1.5em} "reason": "<brief justification>" \\
        \}
        \rmfamily
        \vspace{0.2em}
        
        Output only the JSON object and nothing else.
        
    \end{tcolorbox}
    \caption{The prompt template for the Refresher, handling memory updates and conflict resolution.}
    \label{B.6}
\end{figure*}

\begin{figure*}[t]
    \centering
    \begin{tcolorbox}[
        colback=white,
        colframe=black,
        arc=3mm,
        boxrule=0.8pt,
        width=\linewidth,
        left=4mm, right=4mm, top=4mm, bottom=4mm
    ]
        \textbf{You are an Accuracy Judge.} \\
        Your task is to label a generated answer as \textbf{'CORRECT'} or \textbf{'WRONG'} by comparing it against a ground truth ('gold') answer.
        
        \vspace{0.5em}
        
        \textbf{Input Data:}
        \begin{itemize}[nosep, leftmargin=1.5em]
            \item \textbf{Question ($Q$):} A question posed by one user to another based on prior conversations.
            \item \textbf{Gold Answer ($A_{gold}$):} The concise ground truth (e.g., specific object, date, or event).
            \item \textbf{Generated Answer ($A_{gen}$):} The model's response to be evaluated.
        \end{itemize}
        
        \vspace{0.2em}
        \textbf{Current Instance:} \\
        Question: \texttt{\{question\}} \\
        Gold Answer: \texttt{\{gold\_answer\}} \\
        Generated Answer: \texttt{\{generated\_answer\}}
        
        \vspace{0.5em}

        \textbf{Evaluation Criteria:}
        \begin{enumerate}[label=\arabic*), nosep, leftmargin=*, labelsep=0.5em]
            
            \item \textbf{Semantic Generosity (Topic Matching)} \\
            The generated answer $A_{gen}$ may be verbose. As long as it touches on the \textbf{same core topic or object} as $A_{gold}$ (e.g., ``A shell necklace'' vs. ``I think it was a necklace made of shells from the trip''), mark it as \textbf{CORRECT}.
            
            \item \textbf{Temporal Flexibility} \\
            For time-related questions:
            \begin{itemize}[nosep, leftmargin=1.5em, label=\textbullet]
                \item \textbf{Formats:} Accept different date formats (e.g., ``May 7th'' vs ``7 May'').
                \item \textbf{Relative References:} Accept relative terms (e.g., ``last Tuesday'', ``next month'') if they logically refer to the same time period as $A_{gold}$.
            \end{itemize}
            If the time reference resolves to the same value, mark it as \textbf{CORRECT}.
            
            \item \textbf{Binary Labeling} \\
            You must output exactly one label. Do NOT include both 'CORRECT' and 'WRONG' in the reasoning to avoid parsing errors.
        \end{enumerate}
        
        \vspace{0.5em}

        \textbf{Output Format:} \\
        First, provide a short (one sentence) explanation of your reasoning, then return the label in JSON format:
        
        \ttfamily
        \{ \\
        \hspace*{1.5em} "reasoning": "<one sentence explanation>", \\
        \hspace*{1.5em} "label": "CORRECT" | "WRONG" \\
        \}
        \rmfamily
        \vspace{0.2em}
        
        Output only the JSON object.
        
    \end{tcolorbox}
    \caption{The prompt template for the LLM-as-a-Judge, used to evaluate the factual accuracy of answers.}
    \label{B.7}
\end{figure*}

\end{document}